\documentclass[runningheads,a4paper]{llncs}
\newif\ifdraft

\newcommand{\blue}[1]{\ifdraft{\leavevmode\color{blue}{#1}}\else{\leavevmode\color{black}{#1}}\fi}

\newcommand{\replace}[2]{\ifdraft{\leavevmode\strikeout{#1} \leavevmode\color{blue}{#2}}\else{\leavevmode\color{black}{#2}}\fi}

\newcommand{\strikeout}[1]{\ifdraft{\st{#1}}\else{\vspace{0ex}}\fi}

\usepackage[lined,boxed,linesnumbered]{algorithm2e}

\usepackage{array}

\usepackage{amsmath}
 
\usepackage{cite}
 
\usepackage{endnotes}
 \let\footnote=\endnote
 
\usepackage[english]{babel}
 
\usepackage[T1]{fontenc}

\usepackage[right]{lineno}
 \linenumbers
 
\usepackage{multibib}

\usepackage{multicol} 
 
\usepackage{multirow} 
 
\usepackage[]{natbib} 
 
\usepackage{pifont}
 
\usepackage{rotating} 
 
\usepackage{soul} 
 \sethlcolor{pink}

\usepackage[table]{xcolor}

\usepackage{tipa}
 
\usepackage[utf8]{inputenc} 
 
\usepackage{url} 

\newcommand{\toefl}{\textsc{ToEFL11}}
\newcommand{\efcamdat}{\textsc{EFCamDat2}}
\newcommand{\reddit}{\textsc{REDDIT-L2}}
\newcommand{\reddituk}{\textsc{REDDIT-UK}}
\newcommand{\locness}{\textsc{LOCNESS}}
\newcommand{\toeflbin}{\toefl/\locness}
\newcommand{\efcamdatbin}{\efcamdat/\locness}
\newcommand{\redditbin}{\reddit/\reddituk}

\newcommand{\killpunct}[1]{} 

\newcommand{\side}[1]{\begin{sideways}{#1}\end{sideways}}
 
\newcommand{\PO}{\phantom{0}}

\newcolumntype{.}{D{.}{.}{-1}}

\newcolumntype{M}[1]{>{\centering\arraybackslash}m{#1}}

\newcolumntype{N}{@{}m{0pt}@{}}

\newcolumntype{C}[1]{>{\centering\let\newline\\\arraybackslash\hspace{0pt}}m{#1}}

\newcolumntype{Y}{>{\centering\arraybackslash}X}
 
\begin{document}
\title{Unravelling Interlanguage Facts \\ via \\ Explainable Machine
Learning
}

\titlerunning{Unravelling Interlanguage Facts via Explainable Machine
Learning}

\author{Barbara Berti$^{1}$\thanks{Corresponding author}, 
Andrea Esuli$^{2}$, 
Fabrizio Sebastiani$^{2}$
}

\authorrunning{Barbara Berti, Andrea Esuli, Fabrizio Sebastiani}

\institute{$^{1}$Dipartimento di Lingue e Letterature Straniere \\
Università di Milano \\
20122 Milano, Italy \\
\email{barbara.berti@unimi.it} \\ \mbox{} \\
$^{2}$Istituto di Scienza e Tecnologie dell'Informazione \\
Consiglio Nazionale delle Ricerche \\
56124 Pisa, Italy \\
\email{\{andrea.esuli,fabrizio.sebastiani\}@isti.cnr.it} }

\maketitle

\begin{abstract}
  Native language identification (NLI) is the task of training (via
  supervised machine learning) a classifier that guesses the native
  language of the author of a text. This task has been extensively
  researched in the last decade, and the performance of NLI systems
  has steadily improved over the years. We focus on a different facet
  of the NLI task, i.e., that of analysing the internals of an NLI
  classifier trained by an \emph{explainable} machine learning
  algorithm, in order to obtain explanations of its classification
  decisions, with the ultimate goal of gaining insight into which
  linguistic phenomena ``give a speaker's native language away''. We
  use this perspective in order to tackle both NLI and a (much less
  researched) companion task, i.e., guessing whether a text has been
  written by a native or a non-native speaker. Using three datasets of
  different provenance (two datasets of English learners' essays and a
  dataset of social media posts), we investigate which kind of
  linguistic traits (lexical, morphological, syntactic, and
  statistical) are most effective for solving our two tasks, namely,
  are most indicative of a speaker's L1. We also present two case
  studies, one on Spanish and one on Italian learners of English, in
  which we analyse individual linguistic traits that the classifiers
  have singled out as most important for spotting these L1s. Overall,
  our study shows that the use of explainable machine learning can be
  a valuable tool for the scholar who investigates interlanguage facts
  and language transfer.

\end{abstract}

\section{Introduction}
\label{sec:intro}

\noindent The idea that facts about the acquisition of a second
language (L2) can be learnt by investigating the traces of a speaker's
mother tongue (L1) has been central to research in applied linguistics
for a long time. Already more than 70 years ago,
\citet{fries1945teaching} understood that the interference of a
learner's L1 constituted a major issue in the learning process, and
that comparing the native and the target language was necessary for
theoretical as well as for pedagogical purposes. A few years later,
\citet{lado1957linguistics} endorsed the view that learners of an L2
display a tendency to transfer forms and meanings of their linguistic
and cultural background to the foreign language. \emph{Contrastive
analysis} \citep*{lado1957linguistics, wardhaugh1970contrastive}
centred precisely upon identifying the similarities and differences
between the native and the target language, as well as upon the role
they play in second language acquisition (SLA) processes.  Drawing
upon Corder's (\citeyear{corder1967significance}) research,
\citet{Selinker:1972vt} proposed the notion of \emph{interlanguage}, a
mutable and transitory linguistic system based on rules dissimilar
from the ones characterising either L1 or L2. In Selinker's view, at
every stage of the learning process, the rules governing the
interlanguage are updated in ways that make it unique to each
learner. In this sense, every learner follows a different learning
path.

In general, \emph{language transfer} \citep*{odlin_1989, aarts1998tag,
altenberg2014use, swanbernard2001learn} refers to the idea that,
irrespective of their level of competence, speakers of an L2 have a
tendency to transfer features of their mother tongue to the foreign
language, both in reception and production tasks. Naturally, such
features pertain to all the linguistic subsystems that make up a
speaker's competence, i.e., pragmatics and rhetoric, semantics,
syntax, morphology, phonology, phonetics, and orthography
\citep{odlin2003cross}.

Although one could instinctively liken language transfer to a hurdle
that affects the learning process, its influence is not necessarily
negative. \emph{Negative transfer} (or interference) occurs when L1
and L2 diverge, and the footprint of the former over the latter
generates errors; conversely, when L1 and L2 converge, the learning
process is facilitated, thus leading to \emph{positive transfer}
\citep{schachter1983new, bardovi2018negative}.

Although negative transfer generally results in the production of
errors, it nonetheless represents a functional strategy reflecting the
natural attitude of learners to cope with linguistic challenges and
communicate in spite of the existing gaps
\citep{Jarvis:2012vl}. Indeed, thanks to interpolation and flexibility
in the construction of meaning, the interlocutor can, to some extent,
arrive at making sense of an L1-driven, ill-formed input.

Even though some scholars fail(ed) to recognise the role of language
transfer (e.g., \citep{meisel1981determining, krashen1983newmark}),
the influence exerted by the mother tongue has been vastly
demonstrated by a wealth of studies aimed at analysing learners'
production errors across different educational as well as proficiency
levels (see, amongst others, \citet*{carrio2012contrastive,
kohlmyr2001err, xia2015error, miliander2003we, rosen2006klingt,
ye2004chinese, zhang2010study}). Indeed, such studies have shown that
``even advanced L2 speakers continue to be influenced by their L1 in a
range of domains'' (\citealp[~146]{Gullberg:2011}). Some L1 traces
appear to be indelible and are, therefore, detectable in the
linguistic production.

Naturally, language transfer has been tackled with the qualitative and
quantitative tools of applied linguistics. Studies range from in-depth
analysis of the production of a restricted sample of learners (e.g.,
\citet*{beare2000differences, mu2007investigation}) to the analysis of
specific transfer patterns in large collections of L2 texts (e.g.,
\citet{aijmer2013advances}).

In very much the same vein as corpus studies, we aim to exploit the
wealth of data available from L2 corpora, this time through the
application of techniques from (supervised) \textit{machine learning}
(ML -- \citep{Jordan:2015gc}) and (computational) \textit{native
language identification} (NLI -- \citep{Malmasi:2016tr}). ML is a
branch of computer science that investigates methods for training
algorithms to solve a certain problem. These algorithms learn from
experience, i.e., learn to solve a problem from exposure to instances
of this problem in which the correct solution is known. In ML
approaches to NLI, the problem is that of correctly identifying the L1
of the author of a (spoken or written) text. Of particular interest is
the reasoning that leads the algorithm to choose a particular L1 over
others. The machine's reasoning can, to some extent, be inspected
(using techniques from \emph{explainable machine learning} --
\citep{Belle:2021lk}), thus producing (hopefully new) knowledge on
language transfer.

Indeed, in this paper we aim to show how insight into interlanguage
facts emerging from usage data can be gained through the application
of techniques from explainable ML. We perform computational NLI by
applying a high-accuracy ML algorithm (\emph{support vector machines}
-- SVMs \citep{Zhang:2011fl}) to three publicly available corpora of
English texts in which the L1 of the author (or the nationality of the
author, which we take as a proxy of their L1) is known. Inspecting the
native language identifiers trained by the SVM allows us to determine
which linguistic phenomena the latter deemed the most revealing of the
author's L1. This, in turn, provides the linguist with intuitions
about the transfer-related phenomena that can be detected in these
corpora. We supplement the NLI experiments by additional experiments
on a much less researched companion task, i.e., predicting if a text
has been written by a native or a non-native speaker.

The rest of this paper is organised as follows. In
Section~\ref{sec:MLandNLI}, we introduce, for the benefit of the
non-expert, the machine learning approach to text classification and
to native language identification. The reader who is already familiar
with this approach may skip to Section~\ref{sec:NLIandSLA}, which is
instead devoted to describing our approach to discovering
interlanguage facts that emerge in second-language acquisition by
analysing the parameters of the native language identifier returned by
the machine learning process. In Section~\ref{sec:experiments}, we
describe in detail our experimental setting, including the datasets we
run our experiments on, and our experimental protocol for
investigating both NLI and native vs.\ non-native classification. In
Section~\ref{sec:Results}, we present the results of our experiments,
discussing the accuracy that our classifiers have obtained, analysing
which types of linguistic traits turn out to be most relevant for NLI
and native vs.\ non-native classification, and presenting the
interlanguage facts and the intuitions about language transfer that
emerge from these experiments. Section~\ref{sec:Conclusion} concludes,
pointing at avenues for future research.


\section{Machine learning and native language
identification}\label{sec:MLandNLI}

\noindent NLI belongs to a large family of tasks that collectively go
under the name of computational \textit{authorship analysis} (AA), a
small branch of computer science that investigates methodologies and
techniques for formulating hypotheses regarding the characteristics or
identity of the author(s) of a text of unknown or controversial
paternity. Computational authorship analysis is a discipline with a
fifty-year history (see for example the fundamental study of
\citet{Mosteller:1964gb}), which however has its roots in the
(obviously non-computational) late nineteenth-century pioneering
studies of \citet{Mendenhall:1887hz} and \citet{Lutoslawski:1890os},
who first tackled authorship through quantitative \emph{stylometry}
techniques, according to what, following \citet{Ginzburg:1989wd}, can
be called an ``evidential paradigm''. AA comprises various sub-tasks,
among which
\begin{itemize}

\item authorship verification (AV): given a text and a candidate
  author, determine whether the latter is the author of the former
  \citep{Stamatatos:2016ij};

\item closed-set authorship attribution: given a set of candidate
  authors assumed to contain the true author of the text under study,
  identify that author amongst them \citep{Stamatatos2009:yq};

\item same-author verification: given two texts, determine whether
  they were written by the same author \citep{Koppel:2014bq};

\item author profiling: identify characteristics of the author of a
  text, such as their gender or age group \citep{Tetreault:2012fu}.
  NLI is a special case of author profiling, in which the
  characteristics under study is the L1 of the author.

\end{itemize}
\noindent AA and its sub-tasks have several areas of application,
amongst which cybersecurity (i.e., the prevention of crimes that could
be committed by digital means) and computational forensics (i.e., the
computational analysis of traces of crimes that have already been
committed). Both of these areas of application address contemporary
texts that generally have no cultural value, such as threatening
messages, anonymous letters, or correspondence between
suspects. However, AA has also been applied to literary or historical
texts, proving to be a valuable aid to the work of
philologists.\footnote{The application of AA techniques to
contemporary \emph{oeuvres} can be found in the attempt to identify
the author of the 15th Book of Oz \citep{Binongo:2003fg}, in the
research of Gramscian journalistic texts originally published without
signature \citep{Basile:2008hw}, and in the analysis of the
authenticity of Montale's ``Posthumous Diary'' \citep{Italia:2013cw}.
Examples of applications to ancient texts include the analysis on the
authenticity of Pliny the Younger's ``Letter on Christians'' to Trajan
\citep{Tuccinardi:2017yg}, on the authenticity of Dante Alighieri's
``Epistle to Cangrande'' \citep{Corbara:2019cq}, and others
\citep{Kabala:2020bu,Kestemont:2015lp}.}

Similarly to all other computational AA tasks, NLI rests upon
stylometry, i.e., the quantitative study of the relative frequencies
with which certain linguistic traits are present in the text. Yet,
whilst AA attempts to capture the \emph{stylistic} footprints
unconsciously left by an author, NLI relies on the author's
\emph{L1-related} footprints. It must be pointed out that
computational NLI, as discussed in this paper, does not take into
account the events narrated, the concepts expressed, and/or their
truthfulness or plausibility, and solely analyses linguistic patterns.


\subsection{Native language identification and text
classification}\label{sec:NLIandML}

\noindent NLI is based on ML, the sub-discipline of computer science
that deals with the design of methods for training algorithms to
complete tasks by exposing them to examples in which these tasks were
successfully accomplished. The most important task among those
addressed by ML is data classification. \emph{Classification} is
concerned with assigning a data item to a class chosen from a finite
and predefined set of classes. Classification deals with scenarios in
which such a task is non-deterministic,\footnote{For example,
assigning a natural number to one of the two classes
\textsf{PrimeNumbers} and \textsf{NonprimeNumbers} cannot be
considered a classification problem, since the assignment can be made
deterministically, i.e., without margins of error. Conversely,
assigning a textual comment on a product to one of the two classes
\textsf{Positive} and \textsf{Negative} is a classification problem,
since deciding whether a certain comment conveys a positive or a
negative sentiment requires subjective judgment.} and is based on an
analysis of the content of the data item itself.

NLI can also be formulated in terms of classification, since it
consists of classifying an L2 text into one of $m$ available classes,
with $m$ being the number of possible L1s. Specifically, NLI is an
instance of \emph{text classification}, a task where ML meets
automatic text analysis~\citep{Sebastiani:2002jw}. Automatic text
classification may concern any of several dimensions of the text
(e.g., classification based on the topic the text is about, according
to its literary genre, etc.), all independent of each other. In NLI,
we perform text classification according to the L1 of the author of
the text.

In text classification, a general-purpose learning algorithm (the
\emph{trainer}; in this paper, a support vector machine) ``trains'' an
automatic system (the \emph{classifier}, or \emph{classification
model}; in this paper, a native language identifier) to correctly
assign texts to the classes of interest (which in this paper represent
the possible L1s) by exposing it to a set of texts (the \emph{training
set}) whose true class is revealed to the classifier. Such techniques
are also referred to as \emph{supervised} learning, since, during the
learning phase, the trainer plays the role of a supervisor. In other
words, by examining the training examples, the classifier learns the
linguistic traits that characterise the texts of each class of
interest, and will thus be able to apply this knowledge when asked to
classify previously unseen texts, whose membership in the classes of
interest is unknown. In fact, what the classifier learns is the
statistical correlation between language traits and classes. In
particular, the classifier learns which traits are strongly correlated
with one of the classes (and are thus useful in the classification
process) and which ones do not show any significant correlation with
any of the classes (and are thus of little or no use).


\subsubsection{Linguistic traits and feature
vectors.}\label{sec:LTandFV}

\noindent When building an NLI system, there are two main factors that
must be taken into account owing to the effect they exert on
classification accuracy (i.e., on the ability of the classifier to
guess the right class as frequently as possible): the first concerns
the type of training algorithm to be used (in this paper: a support
vector machine), while the second concerns the language traits that
the algorithm must examine. Whilst the choice of the former is
important, it is probably less so than the choice of the latter. In
fact, while there is a wide range of ML algorithms, and while each of
them displays a different degree of accuracy on a given dataset, it is
a well-known fact that some of these algorithms (among which SVMs)
perform very well in almost all contexts of application.

Conversely, in applications of text classification, it is the choice
of which language traits (``features'', in ML terminology) to base the
analysis upon, that must be carefully pondered. For example, it is
clear that choosing semicolons as a linguistic trait would not be of
much help if we were to perform classification \emph{by topic}, since
the frequency with which punctuation marks are used bears virtually no
relation to the topic of a text. On the contrary, punctuation could be
useful in an NLI task (for the L2s that do use punctuation marks),
because different L1s make use of punctuation in different ways, and
this might interfere with L2 production. Thus, when building an NLI
system, one must choose the linguistic features that, aside from being
easy to analyse algorithmically, one hypothesises to be correlated
with L1 transfer.

\begin{figure}[t]
  \begin{center}
    \includegraphics[width=.60\textwidth]{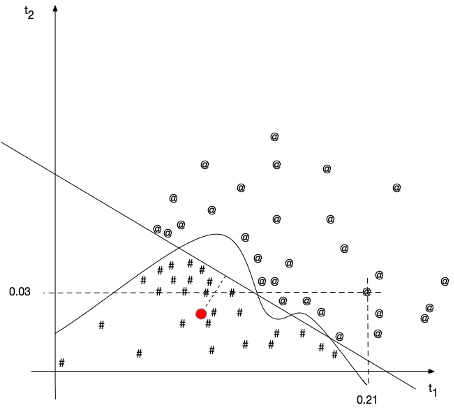}
    \caption{\label{fig:Cartesian}Representation of texts in a
    Cartesian plane.}
  \end{center}
\end{figure}

Once the linguistic features have been chosen, it is possible to
extract from each text a set of relative frequencies of these
features. For each chosen feature, the extraction algorithm will
simply count the occurrences of this feature in the text, divide this
number by the total number of occurrences of any feature, and store
the resulting relative frequency into a data structure called a
\emph{vector}. This is necessary because any data item submitted for
consideration to an ML algorithm must be submitted not in raw form but
in vector form. A vector is an ordered collection of data, in this
case numbers representing relative frequencies of linguistic
features. Each vector representing a text can be viewed as a point in
a Cartesian plane, as in Figure~\ref{fig:Cartesian}.
Each linguistic feature (in Figure~\ref{fig:Cartesian}: $t_{1}$ and
$t_{2}$) corresponds to an axis of the plane, and the relative
frequency of that feature in a text corresponds to the coordinate that
the point representing this text has for that axis. In
Figure~\ref{fig:Cartesian}, the points represent the texts by authors
of two different L1s (L1a and L1b), where the symbols ``\#'' and ``@''
indicate the points corresponding to training texts by L1a authors and
L1b authors, respectively. For ease of illustration, we here assume
that only two linguistic traits ($t_{1}$ and $t_{2}$) are extracted by
the extraction algorithm; in this way, we can generate the familiar
two-dimensional Cartesian plane. For example, for the highlighted
point of type ``@'', the relative frequency of feature $t_{1}$ in the
text is 0.21, whilst the relative frequency of feature $t_{2}$ in the
text is 0.03. Evidently, texts with similar relative frequencies of
occurrence of the same traits are represented by points which are
close to each other in the Cartesian plane. If the linguistic traits
have been chosen well (i.e., if they are good markers of native
language), texts by authors with the same L1 will also be represented
by points that are close to each other in the Cartesian plane.

Figure~\ref{fig:Cartesian} represents a drastic simplification of the
actual NLI process. In fact, tens of thousands of features (instead of
two) are usually considered in a real NLI endeavour; for example, all
words that appear at least once in at least one training document are
usually made to correspond to one feature each. The resulting
\emph{vector space} is thus highly multidimensional, and whilst it can
be treated mathematically on a par with a two-dimensional space, it
cannot be easily displayed in a two-dimensional figure.

With reference to the simple example of Figure~\ref{fig:Cartesian},
for the machine learning algorithm, training an NLI classifier means,
at a first approximation, finding a line in the Cartesian plane that
separates the training examples of L1a from those of L1b; this line
corresponds to the classifier / native language
identifier. Figure~\ref{fig:Cartesian} shows two potential lines that
have this property: a straight and a curved one. Different learning
algorithms choose different lines amongst the many possible ones. In
mathematical terms, any of those lines is identified by (i) a
parametric equation, and (ii) parameter values for this equation. A
learning algorithm is characterised by a certain parametric equation
(e.g., $t_{2}=a\cdot t_{1} +b$, that represents all straight lines in
the Cartesian plane of Figure~\ref{fig:Cartesian}); during the
training phase, it observes the distribution of training examples in
order to determine the parameters of the equation (for the above
equation: its slope and its distance from the origin) so that the
resulting line
best separates the ``\#'' and ``@'' examples.


When a document written by an author of unknown L1 needs to be
classified, the algorithm converts it to a point in the same Cartesian
space (in Figure~\ref{fig:Cartesian}, the point indicated by a small
red dot), using the same conversion process used for the training
documents. Depending on where it is located on the plane, it will end
up either on one side or on the other side of the line that represents
the classifier; this determines whether the classifier decrees it an
L1a text or an L1b text. The distance of this point from the line can
be interpreted as the \emph{degree of certainty} that the classifier
has in determining the class to which the document belongs; a greater
distance corresponds to a greater certainty that the classifier has in
its own classification decision.

In the more general case in which the vector space is $k$-dimensional
(instead of 2-dimensional, as in Figure~\ref{fig:Cartesian}), instead
of a line the classifier is a \textit{hyperplane}, i.e., a surface of
$(k-1)$ dimensions. In the more general case in which there are $n$
possible L1s (instead of 2, as in Figure~\ref{fig:Cartesian}), the
classifier is composed of $(n-1)$ separating surfaces.


\section{Native language identification, second language acquisition,
and explainable machine learning}
\label{sec:NLIandSLA}

\noindent NLI has been investigated fairly extensively in the last
decade. Two main factors have contributed to such increased attention.
The first is the fact that datasets of texts annotated by author's L1,
which could serve as training data and test data for NLI systems, have
become available: these include ICLE \citep{Granger:2009yu}, LANG8
\citep{Brooke:2011cr}, \toefl\ \citep{Blanchard:2013kx}, \efcamdat\
\citep{Geertzen:2013tx}, and \reddit\ \citep{Rabinovich:2018xz}. The
second is the fact that NLI ``shared tasks'' (i.e., evaluation
campaigns) have been
organized~\citep{Tetreault:2013fv,Kumar:2017ii,Kumar:2018fy,Malmasi:2017hk},
and these competitive settings have driven many researchers to develop
increasingly better methods that could measure up with, or beat, the
state of the art.

However, these two factors have mostly pushed researchers to optimize
sheer performance, and have not necessarily incentivised them to
interpret their systems' output in terms of the linguistic phenomena
that underlie NLI. In this respect, it has often been pointed out
\citep{Malmasi:2015fv,Tetreault:2012fu} that one of the potential
outcomes of NLI is the possibility of gaining insight into the
L1-related factors that shape language transfer. Although in recent
years the number of publications in the field of NLI has been growing
in a bid to improve the accuracy of NLI software, a qualitative
post-hoc inspection and further reflection on the results obtained
from an SLA perspective, is lacking. First attempts at exploiting the
insights provided by machine-learning-based NLI to unravel facts about
language transfer were made by \citet{Jarvis:2012vl}
and \citet{Jiang:2014so}, but not many have followed in the same
tradition.

The present work sets out to bridge this gap by analysing the
classifiers produced by the machine learning algorithms, according to
the tenets of \emph{explainable machine learning} (EML -- see e.g.,
\citep{Belle:2021lk}). In the traditional ML approach to text
classification, the classifier produced as the output of a machine
learning process is usually a ``black box'', i.e., a function that
observes a document and assigns to it a class label without providing
any explanation as to the reasons that led to such an assignment. On
the contrary, in EML the classifier that has led to a certain
decision, and the route it has taken to reach it, can be inspected,
making explicit (in human-readable form) the
rules/patterns/correlations that were exploited by the classifier in
order to perform this class label assignment. In this study, we aim to
inspect the algorithm's rules/patterns/correlations in order to gain
insights into the processes at work in SLA.

More specifically, we use a support vector machine to train a
classifier to perform native language identification, using a training
set of texts whose authors' L1s are known. The classifier generated by
an SVM is a vector of parameters, one for each feature. Once the
training phase is completed we inspect the parameters of the
classifier. The numerical value of a parameter is the information that
determines how the value of the corresponding feature's relative
frequency in a document contributes to form the classification
decision for that document. A high absolute value for a parameter
denotes a strong contribution of the feature associated to it in
determining the classification decision (i.e., it indicates that the
SVM believes that this feature has a high discriminative power),
whereas the sign of the value determines if the contribution is toward
choosing a specific L1 or against choosing it. \emph{This means that,
e.g., a feature to which the SVM has associated a positive value of
high magnitude, corresponds to a footprint that speakers of the L1
considered often leave in their L2 production.}

The reliability of the insights that we can thus obtain depend, of
course, on the accuracy of the classifier. The parameters of an
inaccurate classifier \replace{do not really say much}{are of little
use} for our purposes since they do not actually contribute to
\replace{taking}{making} \emph{correct} classification
decisions. \replace{instead}{Conversely,} the parameters of an
accurate classifier \replace{contain}{carry} valuable information,
being the key elements in \replace{taking}{making} correct
classification decisions.

It must be pointed out that the processes at work in SLA need not give
rise to errors. For instance, learners belonging to a certain L1
\blue{community} might be inclined to overuse a legitimate L2
structure \replace{in light of the fact that}{if} it literally
translates a frequently used pattern in their mother tongue. Albeit
\replace{incorrect}{correct}, excessive reliance on \replace{that
particular}{a certain} pattern turns into a distinctive trait for
\replace{that}{a specific} L1 group. At the other end of the spectrum
is \emph{avoidance} \citep{dusk69}, a consequence of L1 and L2
divergence; accordingly, learners tend to steer clear of the
structures that are not typical of their L1, whilst, at the same time,
they rely upon the ones they are familiar with, thus making their L2
production distinctive. Indeed, one aim of our investigation is to
detect patterns of overuse (indicated by a positive value of high
magnitude) and/or underuse (indicated by a negative value of high
magnitude) common to speakers of the same community.

NLI has always been tackled by using corpora consisting of texts
(usually essays) written in a common L2 (usually English) by
\replace{second-language}{L2} learners belonging to many L1 groups. In
machine learning, this corresponds to a \emph{single-label multiclass}
classification task, since each text must be assigned to exactly one
(``single-label'') of $n>2$ (``multiclass'') possible L1s. In this
work we go one step further, and also analyse binary (i.e., $n=2$)
corpora containing texts not necessarily written by
\replace{second-language}{L2} learners. In other words, we will
consider corpora consisting of texts written in a common language (in
our case: English), some of which have been written by native speakers
of this language and some of which have been written by
\replace{second-language}{L2} speakers; while in the multiclass case
the set of classes is, say,
$\{\textsf{Italian}, \textsf{French}, \textsf{Chinese}, ...\}$, in the
binary class it is
$\{\textsf{Native},\textsf{Non-Native}\}$. 
We decided to experiment with corpora containing native texts too in a
bid to extract further discriminant features. The rationale for this
choice is that a comparison between native and non-native texts might
bring to the surface patterns that are not only shared amongst
speakers with the same linguistic background, but that also mark their
output as non-native. In fact, inspection of the processes at work in
a multiclass classification task only provides insight into the
patterns that distinguish a specific L1 from all other L1s, whilst it
does not disclose information about how the output deviates from the
(native) ``norm''. Conversely, by comparing native to non-native
texts, we aim to extract more and different patterns that mark L2
speakers, thus \replace{collecting}{gathering} further knowledge on L2
production.


\section{Experimental analysis}
\label{sec:experiments}


\subsection{The corpora}
\label{sec:corpora}

\noindent In this section, we present in detail the corpora we have
used in order to carry out our experimental analysis, starting
(Section~\ref{sec:multiclass}) from the ones we have used for the
multiclass classification task (i.e., detecting the L1 of the author
of the text), and carrying on (Section~\ref{sec:binary}) with the ones
we have used for the binary task (i.e., detecting whether the author
of the text is or is not a native speaker).


\subsubsection{Multiclass classification.}
\label{sec:multiclass}

\noindent In order to carry out a standard, multiclass NLI task, one
needs a corpus consisting of writings of foreign authors whose L1 is
known and manifest. To this aim, we utilised three publicly available
corpora of English as a foreign language, i.e.,
\toefl\footnote{Downloadable at
\url{https://catalog.ldc.upenn.edu/LDC2014T06}}
\citep{Blanchard:2013kx}, \efcamdat\footnote{Downloadable at
\url{https://corpus.mml.cam.ac.uk/resources/}} \citep{huang,
Geertzen:2013tx}, and \reddit\footnote{Downloadable at
\url{http://cl.haifa.ac.il/projects/L2/}} \citep{Rabinovich:2018xz,
goldin-etal-2018-native}. \toefl\ and \efcamdat\ are learner corpora
consisting of writings produced by learners of English, whilst
\reddit\ is a collection of
posts written in English by non-native Reddit.com users.

\toefl\ (standing for ``Test of English as a Foreign Language -- 11
L1s'') is a publicly available dataset that was compiled in 2013 to
support studies in natural language processing and, in particular, in
NLI. It aims to overcome some shortcomings of its predecessor, i.e.,
ICLE \citep{Granger:2009yu}, namely the uneven distribution of topics
across the various L1s. Indeed, the problem of topic distribution is
particularly relevant in NLI, since a corpus characterised by
\replace{an unbalanced such distribution}{such an unbalanced
distribution} could turn the task into topic identification rather
than L1 detection.\footnote{In other words, a classifier set up to
perform NLI might, when trained and applied on such a dataset, perform
unrealistically well, due to the fact that what it actually recognises
is the topic the text is about, rather than the L1 of its author.}
\toefl\ consists of 12,100 essays written by learners of English from
11 L1s (derived from the learners' nationality) and collected on the
occasion of the TOEFL exam sessions held in different countries
between the years 2006 and 2007. The language families covered are
Romance (French (FRE), Italian (ITA), Spanish (SPA)), Germanic (German
(GER)), Indo-Iranian (Hindi (HIN)), Altaic (Japanese (JPN), Korean
(KOR), Turkish (TUR)), Sino-Tibetan (Chinese (CHI)), Afro-Asiatic
(Arabic (ARA)), and Dravidian (Telugu (TEL)). Each language is
represented by 1,100 essays evenly sampled from eight prompts (see
Table~\ref{tab:TOEFL} from \citep{Blanchard:2013kx}).
As to the length of the essays, on average, it varies between 300 and
350 words.

\newcolumntype{C}{>{\centering\arraybackslash}p{0.06\textwidth}|}

\begin{table}[htbp]
  \caption{Number of documents per language per prompt (all columns
  but last) and total number of tokens per language (last column) in
  the \toefl\ dataset.}
  \begin{center}
    \begin{tabular}{|r||CCCCCCCC|c|}
      \hline
      L1 & P1 & P2 & P3 & P4 & P5 & P6 & P7 & P8 & \# of
                                                   tokens \\
      \hline
      ARA & 138 & 137 & 138 & 139 & 136 & 133 & 138 & 141 &
                                                            309,995 \\
      CHI & 140 & 141 & 126 & 140 & 134 & 141 & 139 & 139 &
                                                            362,176 \\
      FRE & 158 & 160 & 87 & 156 & 160 & 68 & 151 & 160 &
                                                          354,978 \\
      GER & 155 & 154 & 157 & 151 & 150 & 28 & 152 & 153 &
                                                           377,801 \\
      HIN & 161 & 162 & 163 & 86 & 156 & 53 & 158 & 161 &
                                                          385,040 \\
      ITA & 173 & 89 & 138 & 187 & 187 & 12 & 173 & 141 &
                                                          324,793 \\
      JPN & 116 & 142 & 140 & 138 & 138 & 142 & 141 & 143 &
                                                            312,571 \\
      KOR & 140 & 133 & 136 & 128 & 137 & 142 & 141 & 143 &
                                                            336,799 \\
      SPA & 141 & 133 & 54 & 159 & 134 & 157 & 160 & 162 &
                                                           362,720 \\
      TEL & 165 & 166 & 167 & 55 & 169 & 41 & 166 & 171 &
                                                          360,353 \\
      TUR & 169 & 145 & 90 & 170 & 147 & 43 & 167 & 169 &
                                                          352,808 \\
      \hline
    \end{tabular}
  \end{center}
  \label{tab:TOEFL}
\end{table}

\efcamdat\ (standing for ``EF-Cambridge Open Language Database version
2'') is a publicly available 83-million-word collection of writing
tasks submitted to Englishtown (the online school of EF Education
First\footnote{\url{https://www.ef.edu/}}) by about 174,000 learners
from 188 countries and autonomous territories. As is the case with
\toefl, in \efcamdat\ too nationality was used as an approximation of
the learners' L1. The learners span across 16 levels of proficiency,
thus representing the entire range of language proficiency aligned
with common standards such as TOEFL, IELTS, and the Common European
Framework of Reference for languages (CEFR). The writings are mostly
narrative and cover 128 topics, such as ``Introducing yourself by
email'' or ``Writing a movie review''. The length of texts ranges from
very few words to short narratives or articles, the mean being 6
sentences. This makes \efcamdat\ rather similar to \toefl. Unlike
\toefl, however, in \efcamdat\ the distribution of topics is not
balanced, since the corpus was not especially compiled to support NLI
tasks.
Although \efcamdat\ offers a wide range of L1s, for consistency, we
restricted our attention to the 11 L1s \replace{of}{collected in}
\toefl, i.e., Arabic (ARA), Chinese (CHI), French (FRE), German (GER),
Hindi (HIN), Italian (ITA), Japanese (JPN), Korean (KOR), Spanish
(SPA), Russian (RUS), Turkish (TUR). For each of the selected L1s we
randomly sampled 2,000 scripts, in order to have a balanced
distribution of L1 labels, similarly to
\toefl. Table~\ref{tab:efcamdat} reports the distribution of writings
across the 11 L1s.

\newcolumntype{C}{>{\centering\arraybackslash}p{0.15\textwidth}|}

\begin{table}[htbp]
  \caption{Number of documents per language in the \efcamdat\
  dataset. The 1st column indicates the number of documents in the
  original dataset, the 2nd column indicates the number of documents
  in the subsets we use for our experiments, while the 3rd column
  indicates the number of tokens in these subsets.}
  \label{tab:efcamdat}
  \begin{center}
    \begin{tabular}{|r||CCC}
      \hline
      \multirow{2}{*}{L1} & \multicolumn{1}{c|}{\# docs} & \multicolumn{1}{c|}{\# docs} & \multicolumn{1}{c|}{\# tokens} \\
                                     & \multicolumn{1}{c|}{(original)} & \multicolumn{1}{c|}{(our subsets)} & \multicolumn{1}{c|}{(our subsets)} \\ \hline
      \hline
      ARA & \phantom{00}3,562 & 2,000 & 153,007 \\
      CHI & 165,162 & 2,000 & 168,207 \\
      FRE & \PO41,626 & 2,000 & 193,768 \\
      GER & \PO54,597 & 2,000 & 198,447 \\
      HIN & \PO29,569 & 2,000 & 156,097 \\
      ITA & \PO45,249 & 2,000 & 181,974 \\
      JPN & \PO21,374 & 2,000 & 166,133 \\
      KOR & \phantom{00}5,433 & 2,000 & 164,271 \\
      SPA & \phantom{00}8,187 & 2,000 & 189,502 \\
      RUS & \PO70,208 & 2,000 & 184,893 \\
      TUR & \PO14,199 & 2,000 & 154,380 \\
      \hline
    \end{tabular}
  \end{center}
  \label{default}
\end{table}%

The \reddit\ corpus is a publicly available collection of Reddit.com
posts in English. Reddit.com is a social news aggregation, web content
rating, and discussion website which hosts over 450 million users (as
of early 2022). The content is organised into subcategories, also
known as \emph{subreddits}, by area of interest.
As stated above, the \reddit\ corpus differs from the ones discussed
above in that, whilst the previous ones are collections of written
tasks carried out in educational settings, \reddit\ is composed of
short texts produced by non-native speakers of English in a
recreational setting. Moreover, Reddit non-native users are highly
proficient and possess near-native command of the
language. Conversely, the proficiency levels of \toefl\ and \efcamdat\
learners are more diversified.
The selection of the posts for inclusion in the corpus was operated on
the basis of the information available on the users. Only posts from
users whose provenance could be retrieved as a metadatum were
selected. Specifically, the country of origin of a user was extracted
from the country ``flair'', a metadatum that users (optionally)
specify in some European subreddits (e.g., r/europe). For this reason
the \reddit\ corpus mostly addresses European languages. The rationale
of using the country of residence is as for other learner corpora,
i.e.,
in the absence of an explicit specification of the mother tongue of
the speaker, the country of residence is used as a proxy
\replace{of}{for} it.
The size of the entire dataset amounts to 3.8 billion tokens,
resulting from over 250 million sentences produced by approximately
45,000 users. The topics are extremely varied as the corpus spans over
80,000 different subreddits, and are not equally distributed across
languages. We selected the texts associated with the 11 most popular
L1s, i.e., Finnish (FIN), French (FRE), German (GER), Italian (ITA),
Dutch (NED), Norwegian (NOR), Polish (POL), Portuguese (POR), Rumanian
(ROM), Spanish (SPA), Swedish (SWE), randomly sampling 10,000 posts,
among those longer than 300 characters, for each language. See
Table~\ref{tab:reddit} for summary statistics about \reddit.


\newcolumntype{C}{>{\centering\arraybackslash}p{0.15\textwidth}|}

\begin{table}[htbp]
  \caption{Number of documents per language in the \reddit\ dataset
  (all lines except last) and in the \reddituk\ dataset (last line);
  we use the former for the multiclass experiments and the binary
  experiments, and the latter for the binary experiments only. The 1st
  column indicates the number of documents in the original datasets,
  the 2nd column indicates the number of documents in the subsets we
  use for our experiments, while the 3rd column indicates the number
  of tokens in these subsets.}
  \begin{center}
    \begin{tabular}{|r||CCC}
      \hline
      \multirow{2}{*}{L1} & \multicolumn{1}{c|}{\# docs} & \multicolumn{1}{c|}{\# docs} & \multicolumn{1}{c|}{\# tokens} \\
                          & \multicolumn{1}{c|}{(original)} & \multicolumn{1}{c|}{(our subsets)} & \multicolumn{1}{c|}{(our subsets)} \\ \hline
      GER & \PO5,882,569 & 10,000 & 1,430,132 \\
      NED & \PO4,896,785 & 10,000 & 1,395,062 \\
      SWE & \PO3,185,234 & 10,000 & 1,401,137 \\
      FRE & \PO2,253,954 & 10,000 & 1,400,692 \\
      FIN & \PO2,209,668 & 10,000 & 1,451,496 \\
      POL & \PO1,827,281 & 10,000 & 1,410,382 \\
      NOR & \PO1,554,218 & 10,000 & 1,380,917 \\
      SPA & \PO1,399,016 & 10,000 & 1,444,177 \\
      POR & \PO1,374,597 & 10,000 & 1,456,318 \\
      ROM & \PO1,175,844 & 10,000 & 1,488,857 \\
      ITA & \PO1,031,113 & 10,000 & 1,519,165 \\ 
      \hline
      ENG & 13,310,178 & 10,000 & 1,439,863 \\ 
      \hline
    \end{tabular}
  \end{center}
  \label{tab:reddit}
\end{table}%

\subsubsection{Binary classification.}
\label{sec:binary}

\noindent As stated in Section~\ref{sec:NLIandSLA}, aside from the
more traditional NLI task in which only datasets of non-native
speakers are utilised, we set out to investigate the differences
between native vs.\ non-native texts, focusing on native vs.\
non-native speakers of English.

Since there exist no ready-made datasets with the above
characteristics, we create binary datasets of native vs.\ non-native
texts by pairing a non-native dataset (one of those discussed in
Section~\ref{sec:multiclass}) with a native dataset containing texts
of a similar type.
For every L1 in the three non-native datasets, we create also a native
vs.\ non-native binary dataset by pairing the portion of a dataset
relative to the specific L1 with a dataset of native documents.

We pair both \toefl\ and \efcamdat\ with the \locness\ corpus
\citep{Granger:1998cu}.\footnote{Downloadable at
\url{https://www.learnercorpusassociation.org/resources/tools/LOCNESS-corpus/}}
\locness\ is a 324,304 word-long collection of 1,933 argumentative
essays written by English native speakers, i.e., American and British
students. In particular, \locness\ is composed of British pupils'
A-level essays (224 essays, for a total of 60,209 words), British
university students' essays (889 essays, 95,695 words), and American
university students' essays (820 essays, 168,400 words). These
pairings are reasonable, since \toefl\ and \efcamdat\ too are
collections of students' writings produced in an educational
setting. The result is two L1-vs-EN corpora, that we call \toeflbin\
and \efcamdatbin, respectively.

\toeflbin\ is composed of 11 L1-vs-EN binary classification datasets,
each consisting of 1,100 native and 1,100 non-native documents. The
\locness\ corpus contains 1,933 documents; in order to work with
balanced native/non-native datasets we randomly sample 1,100 documents
from it in order to define the native portion of each \toeflbin\
dataset. All the resulting 11 binary datasets use the same sample of
1,100 \locness\ documents.\footnote{We have run repeated experiments
using different samples of 1,100 \locness\ documents without observing
significant variations in the results.}

\efcamdatbin\ is composed of 11 L1-vs-EN binary classification
datasets, each consisting of 1,933 native documents and 2,000
non-native documents. Given the small difference in size between the
native and the non-native portions of \efcamdatbin\ datasets, we have
not performed under-sampling on the majority label, and we consider
the \efcamdatbin\ datasets to be balanced.

Concerning \reddit, we create the native vs.\ non-native datasets
using \textsc{REDDIT-UK}, an addendum to the \reddit\ corpus that
\replace{collects}{comprises} Reddit.com posts produced by native
British English speakers only. We call \redditbin\ the resulting
L1-vs-EN dataset; it is composed of 11 L1-vs-EN binary classification
datasets, each consisting of 10,000 native and 10,000 non-native
documents. All the resulting 11 binary datasets use the same sample of
1,100 \textsc{REDDIT-UK} documents.

Table~\ref{tab:datasets} summarizes the characteristics of all the
datasets we use in our experimentation.

\newcolumntype{C}{>{\centering\arraybackslash}p{0.15\textwidth}|}

\begin{table}[htbp]
  \caption{Corpora used in this work, and their characteristics. All
  these datasets contain documents in 11 L1 languages.}
  \begin{center}
    \begin{tabular}{|r||c|CC}
      \hline
      Dataset & Type & \# of texts & \# of tokens \\
      \hline\hline
      \toefl\ & Non-native & \PO12,100 & \PO3,840,034
      \\
      \efcamdat\ & Non-native & \PO22,000 & \PO1,910,679 \\
      \reddit\ & Non-native & 110,000 & 15,778,335
      \\
      \hline
      \toeflbin\ & Native vs.\ Non-native & \PO13,200 & \PO4,026,257 \\
      \efcamdatbin\ & Native vs.\ Non-native & \PO23,933 & \PO2,234,983\\
      \redditbin\ & Native vs.\ Non-native & 120,000 & 17,218,198\\
      \hline
    \end{tabular}
  \end{center}
  \label{tab:datasets}
\end{table}%


\subsection{Features}
\label{sec:features}

\noindent
We decided to examine the contribution of different types of features
to the NLI endeavour: lexical, morphological, syntactic, and
statistical.


\subsubsection{Lexical features.}
\label{sec:lexicalfeatures}

\noindent We start by considering as lexical features the tokens
(i.e., words as they appear in the text, which we call ``type T''
features) or the lemmas (i.e., every token is reduced to its
corresponding lemma, giving rise to what we call ``type L''
features).\footnote{By ``tokens'' we mean individual words, including
function words and punctuation symbols; a punctuation symbol generates
a distinct token even when it is attached to a word. In order to
perform tokenisation we use the SpaCy tool (\url{https://spacy.io/}),
which we also use to perform all the natural language processing other
than tokenisation and lemmatisation mentioned in the rest of the
paper, i.e., sentence splitting, part-of-speech (POS) tagging,
named-entity recognition, extraction of morphological suffixes,
dependency parsing.} In both cases we consider unigrams, bigram, and
trigrams (i.e., sequences of one / two / three tokens / lemmas), thus
generating the six sets of features T1, T2, T3, L1, L2, L3.

We also test a ``masked'' version of the above features in which named
entities (NE) are replaced by a placeholder.\footnote{In order to
extract named entities we use the SpaCy named-entity recognition model
``en\_core\_web\_md'' (\url{https://spacy.io/models/en}), which is
reported to have a very good macro-$F_1$ score (macro-$F_1$ being an
accuracy measure, with 0 representing minimum accuracy and 1
representing maximum accuracy) of 0.84 on a set of 18 entity labels.}
NEs denote real-world objects such as organisations, locations,
persons, etc., and represent an issue in NLI, since they are clues
that the classifier could heavily rely upon in order to classify the
texts. For instance, \textsc{ToEFL} / \efcamdat\ assignments that
prompt the candidates to describe personal habits are likely to favour
the use of terms such as geographical locations (e.g., Italy, Tuscany,
Rome, etc.), languages (e.g., Finnish, Norwegian, Swedish, etc.),
proper nouns (e.g., Javier, Pilar, Rocío, etc.), organisations (e.g.,
Peugeot, Sorbonne, Paris Saint-Germain, etc.), currencies (e.g., Yuan,
Yen, Ruble, etc.), and so on. As a consequence, the classifier could
end up assigning the correct label to a document
\replace{just}{solely} in virtue of the NEs it contains. For this
reason, we define sets of features from which NEs are masked out (we
call the resulting feature sets TN1, TN2, TN3, LN1, LN2, LN3).

We also test the use of a different form of masking that masks out all
terms belonging to some specific POS classes; the POS classes that are
masked are ADD (email address), FW (foreign word), NN (noun, singular
or mass), NNP (noun, proper singular), NNPS (noun, proper plural), NNS
(noun, plural), XX
(unknown). 
The masked terms are replaced with their POS tags (e.g., ``Reach me at
john.doe@gmail.com'' becomes ``Reach me at ADD''). We call the
resulting feature sets TP1, TP2, TP3, LP1, LP2, LP3.


\subsubsection{Morphological features.}
\label{sec:morphologicalfeatures}

\noindent We also study the role of morphological suffixes, by mapping
tokens into pairs consisting of a POS tag and a morphological
suffix. Such a mapping would transform, e.g., the sentence ``the
election of the president is heating quickly'' into the sequence ``DET
NN-ction IN DET NN-ent VB VB-ing RB-ly'', from which features could be
extracted as usual (e.g., ``DET'' would be a unigram and ``DET
NN-ction'' would be a bigram). We call the resulting feature sets MS1,
MS2, MS3. The hypothesis we want to test here is that speakers of
different L1s might have a tendency to choose different English terms
based on their morphological similarity with terms in their respective
L1s.

\subsubsection{Syntactic features.}
\label{sec:syntacticfeatures}

\noindent As for the syntactic part, we map all the words in the text
into
\begin{itemize}

\item their respective parts of speech (e.g., ``I run fast'' becomes
  ``PRP VBP RB''); this gives rise to the P1, P2, P3 feature sets;

\item the respective labels obtained from syntactic dependency
  parsing, which assigns a syntactic label to each token (e.g.,
  ``adverbial modifier'', ``clausal subject'', etc.); this gives rise
  to the D1, D2, D3 feature sets. The rationale behind our use of
  syntactic parsing is that speakers of different L1s may structure
  their sentences differently, and we may thus expect syntactic
  parsing to capture these habits.
\end{itemize}


\subsubsection{Statistical features.}
\label{sec:statisticalfeatures}

\noindent Finally, we define three sets of statistical features, by
analysing word lengths (WL), sentence lengths (SL), and dependency
depths (DD).

Analysing word lengths means mapping a text into a list of numbers
that denote the lengths of the words that make up the text (e.g., ``I
have lived in France all my life'' would be encoded as ``1 4 5 2 6 3 2
4'', which can be represented by features 1 2 3 4 5 6).

Analysing sentence lengths is similar, but is performed at the
sentence level, i.e., means mapping a text into a list of numbers that
denote the lengths of the sentences that make up the text.

Analysing dependency depths means \replace{instead}{} measuring the
number of hops that are necessary to ``jump'' from the root of the
dependency parse tree to the node of the tree that represents the
specific token. For instance ``I like cookies that contain butter''
would be encoded as ``1 0 1 3 2 3'', given that ``like'' is the root
verb, ``I'' and ``cookies'' are directly linked to it, ``contain'' is
linked to ``cookies'', and ``that'' and ``butter'' are linked to
``contain''.  Dependency depths are correlated with sentence length,
but provide specific information concerning syntactic complexity.

\bigskip

After extracting all these features, we filter out all lexical,
morphological, and syntactic features that occur only once in a given
dataset, since these features cannot possibly have an impact on the
classification process (if a feature occurs in the training set but
not in the test set, no test document will be impacted by it; if a
feature occurs in the test set but not in the training set, no
knowledge about its correlation with the class labels has been gained
during the training phase).

Table~\ref{tab:feat_count} reports how many distinct features of each
type remain in each dataset after the above-mentioned filtering step.
Features of type 2 and 3 (bigrams and trigrams) show a combinatorial
growth in number with respect to features of type 1 (unigrams), as
expected. \replace{Part-of-speech} {Conversely, part-of-speech} (P1),
dependency-depth (D1), and statistical features (WL SL DD), are few.
We cannot expect these small-sized feature sets to bring about a high
classification accuracy by themselves, but it will be interesting to
inspect which features from these sets are the most informative for
the machine learning algorithm.



\newcolumntype{C}{>{\raggedleft\arraybackslash}p{0.11\textwidth}|}

\begin{table}[htbp]
  \caption{Number of features extracted from each dataset.}
  \label{tab:feat_count}
  \begin{center}
    \begin{tabular}{|c|c|CCCCCC}
      \cline{3-8} 
      \multicolumn{2}{c|}{} & \multicolumn{1}{c|}{\side{\toefl}} & \multicolumn{1}{c|}{\side{\efcamdat}} & \multicolumn{1}{c|}{\side{\reddit}} 
      & \multicolumn{1}{c|}{\side{\toeflbin}} & \multicolumn{1}{c|}{\side{\efcamdatbin}} & \multicolumn{1}{c|}{\side{\redditbin\PO}} \\
      \hline
      \multirow{18}{*}{Lexical}
                            & T1 & 25,074 & 19,369 & 104,721 & 28,550 & 23,946 & 109,575 \\ 
                            & T2 & 186,418 & 106,775 & 814,017 & 209,404 & 135,311 & 873,251 \\ 
                            & T3 & 328,578 & 159,226 & 1,194,267 & 358,245 & 188,701 & 1,301,792 \\ 
                            & L1 & 19,542 & 15,722 & 88,901 & 22,255 & 19,097 & 93,224 \\ 
                            & L2 & 148,110 & 89,016 & 681,563 & 167,523 & 113,222 & 729,723 \\ 
                            & L3 & 308,919 & 149,001 & 1,140,693 & 338,838 & 179,960 & 1,240,682 \\ 
                            & TN1 & 23,712 & 16,328 & 85,183 & 26,576 & 20,387 & 89,021 \\ 
                            & TN2 & 184,608 & 99,357 & 749,645 & 206,224 & 126,524 & 802,801 \\ 
                            & TN3 & 330,170 & 155,540 & 1,217,103 & 360,332 & 185,426 & 1,325,138 \\ 
                            & LN1 & 18,189 & 12,606 & 68,676 & 20,264 & 15,425 & 71,923 \\ 
                            & LN2 & 146,124 & 81,231 & 612,688 & 164,051 & 104,006 & 654,319 \\ 
                            & LN3 & 310,343 & 144,542 & 1,154,485 & 340,745 & 175,948 & 1,253,926 \\ 
                            & TP1 & 9,187 & 6,074 & 28,412 & 10,223 & 7,596 & 29,832 \\ 
                            & TP2 & 79,184 & 43,886 & 282,027 & 87,483 & 55,312 & 299,864 \\ 
                            & TP3 & 210,196 & 102,120 & 794,618 & 229,087 & 124,785 & 854,377 \\ 
                            & LP1 & 5,928 & 3,986 & 17,742 & 6,502 & 4,768 & 18,681 \\ 
                            & LP2 & 53,946 & 30,717 & 190,470 & 59,676 & 38,737 & 201,825 \\ 
                            & LP3 & 172,577 & 83,354 & 644,229 & 188,877 & 103,865 & 689,800 \\ 
      \hline
      \multirow{3}{*}{Morphological}
                            & MS1 & 9,335 & 6,199 & 28,600 & 10,376 & 7,743 & 30,021 \\ 
                            & MS2 & 100,182 & 54,326 & 334,674 & 110,281 & 68,357 & 355,830 \\ 
                            & MS3 & 285,928 & 130,648 & 1,016,853 & 311,470 & 159,616 & 1,096,187 \\ 
      \hline
      \multirow{6}{*}{Syntactic}
                            & P1 & 50 & 50 & 50 & 50 & 50 & 50 \\ 
                            & P2 & 1,677 & 1,656 & 2,160 & 1,754 & 1,751 & 2,174 \\ 
                            & P3 & 18,417 & 16,144 & 42,420 & 20,437 & 18,567 & 43,341 \\ 
                            & D1 & 45 & 45 & 45 & 45 & 45 & 45 \\ 
                            & D2 & 1,534 & 1,402 & 1,802 & 1,553 & 1,465 & 1,814 \\ 
                            & D3 & 16,658 & 12,712 & 29,815 & 17,409 & 14,301 & 30,542 \\ 
      \hline
      \multirow{3}{*}{Statistical}
                            & WL & 21 & 38 & 41 & 28 & 42 & 39 \\ 
                            & SL & 159 & 111 & 192 & 161 & 115 & 196 \\ 
                            & DD & 28 & 21 & 65 & 29 & 29 & 65 \\ 
      \hline
    \end{tabular}
  \end{center}
  \label{tab:features}
\end{table}%


\subsection{Experimental protocol}
\label{sec:protocol}

\noindent We run experiments of two types, i.e., (i) multiclass
experiments on \toefl, \efcamdat, and \reddit, aimed at determining
the L1 of the author of the text, and (ii) binary experiments on
\toeflbin, \efcamdatbin, and \redditbin, aimed at determining whether
the author of the text is a native or non-native speaker of English.

As previously mentioned, the learning algorithm we use for our
experiments is support vector machines (SVMs)\footnote{In some
preliminary experiments we alternatively tested the use of
decision-tree and decision-forest learning algorithms, but we found
the resulting classifiers difficult to inspect in an NLI scenario, as
the very high number of features produces very complex and deep trees
that do not clearly show interpretable patterns. We use the
implementation of linear SVMs from the \texttt{scikit-learn}
Python-based package
(\url{https://scikit-learn.org/stable/index.html}). All the code that
allows to replicate the experiments is available at
\url{https://github.com/aesuli/nli-exp22}.}.
SVMs were devised for training binary classifiers, so they are
natively fit for running the binary experiments. For the multiclass
experiments, though, \replace{we need a workaround, which consists of
adopting the well-known ``one-vs-all'' approach}{a workaround, i.e.,
the well-known ``one-vs-all'' approach, was necessary}. This approach
comes down to training one binary classifier for each of the $n$ L1s
considered; for each such classifier, the training documents written
by speakers of the L1 considered are used as positive training
examples, while the training documents written by speakers of all the
other $(n-1)$ L1s are used as negative training examples. We then
independently ``calibrate'' each of the resulting $n$ binary
classifiers (via ``Platt calibration'' -- see \citep{Platt:2000fk}),
i.e., we tune them in such a way (i) that each of them outputs, for a
test document, a posterior probability (representing the probability
that the classifier subjectively assigns to the fact that the document
was written by a speaker of the corresponding L1), and (ii) that the
posterior probabilities returned by the $n$ classifiers for the same
test document are comparable.
The set of $n$ calibrated binary classifiers can thus be used to
perform multiclass classification of a test document, by (a)
classifying the document with all the $n$ classifiers, and (b)
assigning as its predicted L1 the one associated with the classifier
that has returned the highest classification score. We can then
inspect each binary classifier, using the methodology described in
Section~\ref{sec:NLIandSLA}, so as to determine which features give
the strongest contribution towards assigning the L1 label and which
features give the strongest contribution towards not assigning it,
independently for every L1.

For the native vs.\ non-native binary classification experiments,
instead, we use the SVM to train a single L1-vs-EN classifier for
every L1. We do not compare a L1-vs-EN classifier for a given L1
against the analogous classifiers for the other L1s, as this
classification task focuses on each L1 independently of the others.

For training our classifiers we use the default values for the SVM
hyperparameters (in particular, we use the linear kernel), for three
reasons: (a) in preliminary experiments \replace{that we had run}{},
explicitly optimising these hyperparameters returned only very
marginal improvements; (b) hyperparameter optimisation
\replace{essentially}{} does not impact on the values assigned to the
most informative features, which are the goal of our study, but on the
long tail of the least informative ones; (c) given the many
combinations we test, our experiments are computationally expensive,
and engaging in hyperparameter optimisation would make them
unmanageable.  Concerning computational cost, we stress that, despite
the enormous amount of features at play (e.g., more than 11 million
features are used in the ``All'' experiment of
Table~\ref{tab:results_L1_sets} for \redditbin), we have performed no
feature selection, in order not to remove any information that might
prove interesting in the analysis we will carry out in
Section~\ref{sec:Results}.

This difference among the multiclass tasks and the binary tasks
highlights the difference in the insights one can derive from the
inspection of the classification models. The features in the
multiclass classification models are meant to separate an L1 from the
other L1s, while the features in the binary L1-vs-EN classification
models are meant to separate a specific L1 from native English.



As the mathematical function for measuring the quality of our
classifiers we use the so-called ``vanilla accuracy'' (hereafter:
accuracy) measure, which is simply defined as the fraction of all
classification decisions that are correct. More formally,
\begin{align}
  \label{eq:accuracy}A=\frac{\sum_{\lambda_{i}\in 
  \mathcal{L}}C_{ii}}{\sum_{\lambda_{i},\lambda_{j}\in 
  \mathcal{L}}C_{ij}}
\end{align}
\noindent where $\mathcal{L}$ is the set of languages considered in
the classification task (11 languages in our multiclass experiments
and 2 languages in our native vs.\ non-native experiments) and
$C_{ij}$ represents the number of documents that the classifier
assigned to $\lambda_{i}$ and whose true label is
$\lambda_{j}$. Accuracy values range from 0 (worst) to 1 (best).

In order to determine the accuracy of the classifiers we adopt a
10-fold cross-validation protocol. A $k$-fold cross validation
protocol evaluates the accuracy of a machine learning algorithm by
running multiple experiments on a given dataset. The dataset is split
into $k$ subsets of the same size. A single experiment consists of
training the learning algorithm on $(k-1)$ subsets and testing the
trained model on the remaining subset. This step is repeated $k$
times, every time using a different subset for testing. Once the $k$
experiments are completed, all the texts in the dataset have been
tested upon, yet in a way that correctly excludes them from the
training set. The collected predictions are compared with the true
labels from the dataset, and accuracy can thus be computed. The
cross-validation protocol thus exploits the entire dataset in order to
evaluate a machine learning method, differently from a simpler
train-and-test protocol in which only a subset of the dataset is
subject to evaluation.

Consistently with the rest of the NLI literature, we use tfidf
weighting for generating all the vectors that represent our documents.


\section{Results}
\label{sec:Results}


\subsection{Results of the multiclass experiments: Identifying the NLI
of the speaker}
\label{sec:NLIresults}

\begin{figure}[!t]
  \includegraphics[width=\textwidth]{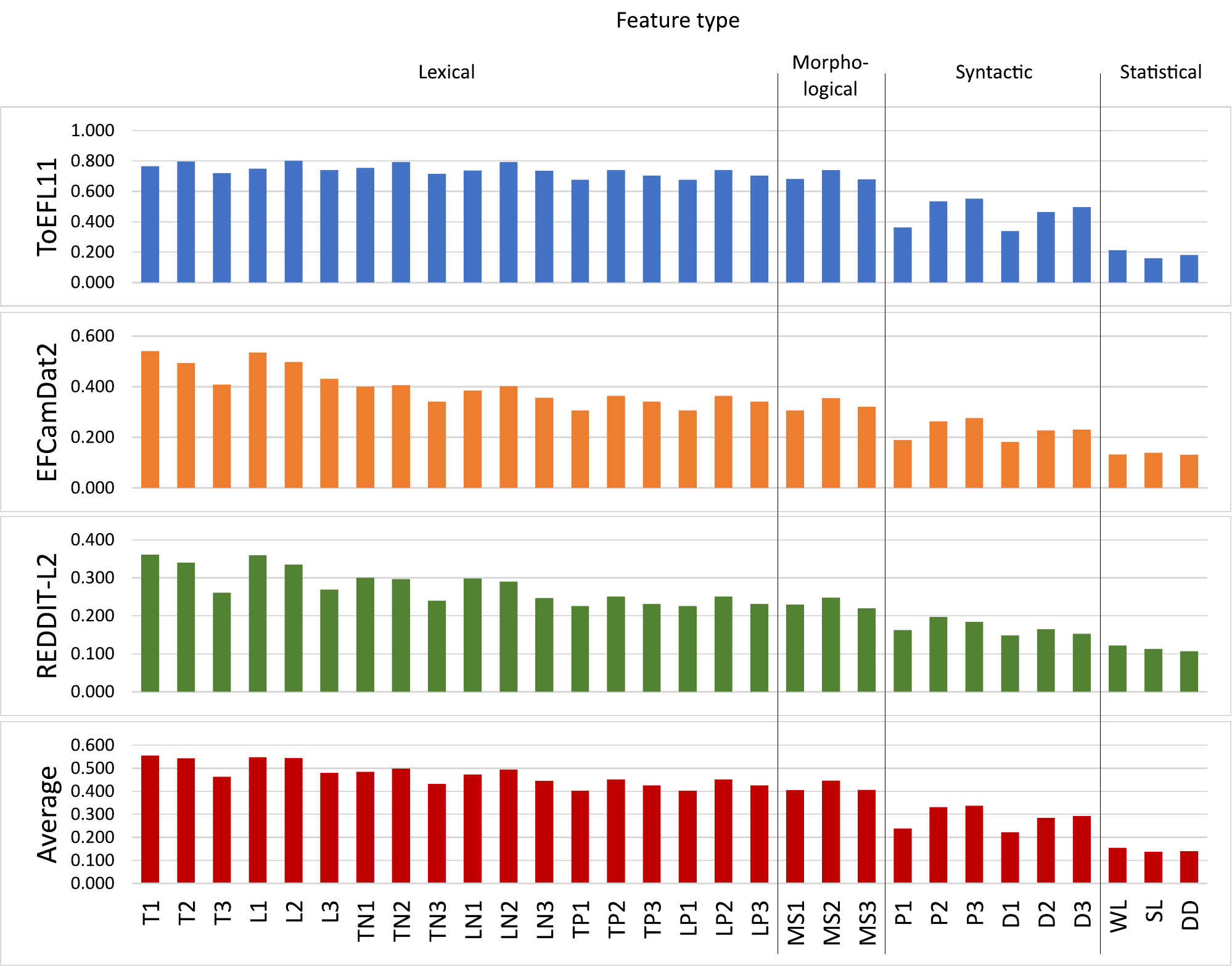}
  \caption{\label{fig:results_L1}Visual comparison among the accuracy
  values deriving from the various feature sets for the NLI
  multi-class classification task. The exact values are reported in
  Table~\ref{tab:results_L1}.}
\end{figure}

\newcolumntype{C}{>{\centering\arraybackslash}p{0.17\textwidth}|}

\begin{table}[htbp]
  \caption{Accuracy results for the L1 identification task, obtained
  by using only the features indicated on the row on the dataset
  indicated in the column. The five best results for every dataset are
  highlighted in \textbf{bold}, and the five worst results are
  highlighted in \textit{italic}. A visual comparison of the values is
  shown in Figure~\ref{fig:results_L1}.}
  \label{tab:results_L1}
  \begin{center}
    \begin{tabular}{|c|c|CCC|c|}
      \cline{3-6}
      \multicolumn{2}{c|}{} & \multicolumn{1}{c|}{\toefl} & \multicolumn{1}{c|}{\efcamdat} & \multicolumn{1}{c||}{\reddit} & \multicolumn{1}{c|}{Average} \\
      \hline
      \multirow{18}{*}{Lexical}
                            & T1 & \textbf{0.766} & \textbf{0.541} & \textbf{0.361} & \textbf{0.654} \\
                            & T2 & \textbf{0.797} & \textbf{0.493} & \textbf{0.340} & \textbf{0.645} \\
                            & T3 & 0.721 & 0.408 & 0.261 & 0.565 \\
                            & L1 & 0.750 & \textbf{0.535} & \textbf{0.360} & \textbf{0.643} \\
                            & L2 & \textbf{0.801} & \textbf{0.497} & \textbf{0.335} & \textbf{0.649} \\
                            & L3 & 0.741 & \textbf{0.431} & 0.269 & 0.586 \\
                            & TN1 & 0.754 & 0.400 & \textbf{0.300} & 0.577 \\
                            & TN2 & \textbf{0.793} & 0.406 & 0.297 & \textbf{0.600} \\
                            & TN3 & 0.716 & 0.341 & 0.240 & 0.529 \\
                            & LN1 & 0.738 & 0.385 & 0.298 & 0.562 \\
                            & LN2 & \textbf{0.793} & 0.402 & 0.290 & 0.598 \\
                            & LN3 & 0.735 & 0.356 & 0.247 & 0.546 \\
                            & TP1 & 0.676 & 0.306 & 0.226 & 0.491 \\
                            & TP2 & 0.741 & 0.363 & 0.251 & 0.552 \\
                            & TP3 & 0.703 & 0.341 & 0.231 & 0.522 \\
                            & LP1 & 0.676 & 0.306 & 0.226 & 0.491 \\
                            & LP2 & 0.741 & 0.363 & 0.251 & 0.552 \\
                            & LP3 & 0.703 & 0.341 & 0.231 & 0.522 \\
      \hline
      \multirow{3}{*}{Morphological}
                            & MS1 & 0.681 & 0.306 & 0.230 & 0.494 \\
                            & MS2 & 0.739 & 0.354 & 0.248 & 0.547 \\
                            & MS3 & 0.679 & 0.321 & 0.220 & 0.500 \\
      \hline
      \multirow{6}{*}{Syntactic}
                            & P1 & \textit{0.363} & \textit{0.189} & 0.162 & \textit{0.276} \\
                            & P2 & 0.535 & 0.263 & 0.197 & 0.399 \\
                            & P3 & 0.551 & 0.276 & 0.184 & 0.414 \\
                            & D1 & \textit{0.339} & \textit{0.182} & \textit{0.148} & \textit{0.261} \\
                            & D2 & 0.464 & 0.227 & 0.165 & 0.346 \\
                            & D3 & 0.495 & 0.231 & \textit{0.153} & 0.363 \\
      \hline
      \multirow{3}{*}{Statistical}
                            & WL & \textit{0.211} & \textit{0.132} & \textit{0.122} & \textit{0.172} \\
                            & SL & \textit{0.160} & \textit{0.139} & \textit{0.113} & \textit{0.150} \\
                            & DD & \textit{0.181} & \textit{0.131} & \textit{0.107} & \textit{0.156} \\
      \hline
    \end{tabular}
  \end{center}
\end{table}%

\newcolumntype{C}{>{\centering\arraybackslash}p{0.17\textwidth}|}

\begin{table}[htbp]
  \caption{Accuracy results for the L1 identification task, tackled by
  using unions of sets of features. The two best results for every
  dataset are displayed in \textbf{bold}, and the two worst results
  are displayed in \textit{italic}.}
  \label{tab:results_L1_sets}
  \begin{center}
    \begin{tabular}{|c|c|CCC|c|}
      \cline{3-6}
      \multicolumn{2}{c|}{} & \multicolumn{1}{c|}{\toefl} & \multicolumn{1}{c|}{\efcamdat} & \multicolumn{1}{c||}{\reddit} & \multicolumn{1}{c|}{Average} \\
      \hline
      \multirow{6}{*}{Lexical}
        & T1 T2 T3 & \textbf{0.816} & \textbf{0.548} & \textbf{0.397} & \textbf{0.682} \\
        & L1 L2 L3 & \textbf{0.817} & \textbf{0.552} & \textbf{0.394} & \textbf{0.685} \\
        & TN1 TN2 TN3 & 0.809 & 0.447 & 0.342 & 0.628 \\
        & LN1 LN2 LN3 & 0.812 & 0.446 & 0.338 & 0.629 \\
        & TP1 TP2 TP3 & 0.759 & 0.392 & 0.277 & 0.576 \\
        & LP1 LP2 LP3 & 0.759 & 0.392 & 0.277 & 0.576 \\
      \hline
      \multirow{1}{*}{Morphological}
        & MS1 MS2 MS3 & 0.762 & 0.388 & 0.278 & 0.575 \\
      \hline
      \multirow{2}{*}{Syntactic}
        & P1 P2 P3 & 0.583 & 0.290 & 0.191 & 0.437 \\
        & D1 D2 D3 & \textit{0.518} & \textit{0.242} & \textit{0.157} & \textit{0.380} \\
      \hline
      \multirow{1}{*}{Statistical}
        & WL SL DD & \textit{0.256} & \textit{0.151} & \textit{0.129} & \textit{0.204} \\
      \hline\hline
      \multirow{1}{*}{(All)}
        & (All) & 0.813 & 0.541 & 0.390 & 0.677 \\
      \hline
    \end{tabular}
  \end{center}
\end{table}%

\noindent Table~\ref{tab:results_L1} reports the accuracy results we
have obtained for the L1 identification task on our three multiclass
datasets. Results are displayed per feature set, since we have run
classification experiments in which only one of the 30 sets of
features we have defined in Section~\ref{sec:features} has been used,
so as to highlight which types of features work best.
%
%
As can be observed, even though accuracy differs considerably across
the three datasets, the trends are rather similar \replace{throughout
them}{}, i.e., if feature set $x$ works better than feature set $y$ in
a dataset, the same tends to happen in the two other datasets too. As
a general observation, lexical features perform better if compared to
other types of features, whilst the features associated to the worst
performance are the statistical ones (i.e., WL, SL, DD), whose figures
are indeed very similar across the datasets; but let us analyse this
in more detail.

In \toefl\ lexical features perform extremely well, particularly token
and lemma bigrams. Masking the NEs does not have a significant effect,
and this is obviously due to the fact that, since \toefl\ was
especially created for NLI tasks, NEs had already been removed
\replace{from it}{} by its creators.\footnote{``This preprocessing was
fairly aggressive and expunged both named entities and most other
capitalized words, replacing them with special
tags.''\citep[p. 4]{Blanchard:2013kx}}
In the \efcamdat\ dataset too, lexical features outperform other types
of features, but this time it is unigrams that lead to the highest
accuracy. Interestingly, masking NEs leads to poor performance. This
might be a consequence of the type of essays the dataset consists
of. In fact, \efcamdat\ \replace{consists of}{is made up of} writing
assignments which, amongst others, prompt the author to write about
their life, habits, and so on. Therefore, resorting to NEs might be
quite common practice, and, \replace{if tis is}{should this be} the
case, \replace{it is intuitive that}{}removing them from the texts
results in decreased performance. In addition, the importance of NEs
is also a possible explanation of the remarkable performance of
unigrams (since NEs often consist of one word only).

Accuracy in the \reddit\ dataset is much lower than in \toefl\ and (to
a lesser degree) \efcamdat, and the differences amongst the various
types of features are less substantial than for the two other
datasets. The fact that the native language identifier struggles with
the \reddit\ posts might be due to two factors: (a) text length,
and/or (b) the proficiency level of Reddit.com authors. First,
Reddit.com posts tend to be brief, and this might provide the machine
with too few significant patterns. The other possible reason has to
do, as mentioned above, with the level of proficiency in English of
non-native speakers who are active users of the social medium. Whilst
the \toefl\ and the \efcamdat\ datasets rely on the production of
\emph{learners} of English, Reddit.com users are generally fluent in
English and interact naturally with their peers. If the overall
English level is high, it might be harder to find discriminant
features that markedly separate L1 groups, as the L2 production will
be quite homogeneous and close to that of native authors.

It is also interesting to compare the performance levels delivered by
unigrams, bigrams, and trigrams, respectively.
Figure~\ref{fig:results_L1} makes it visually apparent that, when it
comes to lexical features, unigrams perform better than bigrams and
much better than trigrams; this indicates that learners from different
L1 groups differ in their choice of words (which is intuitive), rather
than in their choice of word groups. The opposite can be observed
\replace{when it comes to}{for} syntactic features, where trigrams
work better than bigrams and much better than unigrams; this indeed
makes sense, since it indicates that different L1 groups differ in
their preferred syntactic constructions, rather than in using one part
of speech more \blue{often} than another.

Table~\ref{tab:results_L1_sets} displays classification accuracy
values averaged across unigrams, bigrams, and trigrams of the same
type; we display the results in this way in order to highlight the
different contributions of the different types of features,
\replace{irrespectively}{irrespective} of the size of the $n$-gram. In
this table, the ``All'' setup refers to an experiment in which
\blue{all} the features are used \blue{simultaneously} \strikeout{all
together}.\footnote{Using \replace{the features all together}{all the
features at once} might give rise to unwanted interactions, since the
same feature might belong \replace{to}{in} more than one group at the
same time (e.g., article ``the'' belongs \replace{to}{in} T1, L1, TN1,
LN1, ..., at the same time. In order to remove unwanted effects
\blue{,} we prefix each feature with the corresponding feature type,
so that, e.g., T1-the, L1-the, TN1-the, LN1-the, ..., all count as
different features.} It is immediately evident from this table that,
in all the three datasets, \textit{the most discriminative features
for the NLI task are the lexical ones, followed by the morphological,
syntactic, and statistical features, in this order}. The lexical
features are so dominant that the two best types (the T and L types)
even deliver better performance than all \blue{the} features
\replace{put}{taken} together; this is somehow unusual for SVMs, which
are notoriously so robust to overfitting that, in general, ``the more
\replace{feature}{features,} the better''. This fact unequivocally
shows that \emph{word choice is, more than anything, what gives an L1
speaker away}.

\begin{table}[htbp]
  \caption{Accuracy results for the L1-vs-EN binary classification
  task on \toeflbin. Each cell represents the accuracy obtained when
  the L1 is the one on the column, using just the features indicated
  on the row.}
  \label{tab:results_bin_toe}
  \begin{center}
    \resizebox{\textwidth}{!} {
    \begin{tabular}{|c|c|c|c|c|c|c|c|c|c|c|c|c||c|}
      \cline{3-14}
      \multicolumn{2}{c|}{} & ARA & CHI & FRE & GER & HIN & ITA & JPN & KOR & SPA & TEL & TUR & Average \\\hline
      \multirow{18}{*}{Lexical}
        & T1 & 0.992 & 0.997 & 0.997 & 0.998 & 0.997 & 0.998 & 0.999 & 0.997 & 0.997 & 0.999 & 0.995 & 0.997 \\
        & T2 & 0.988 & 0.996 & 0.992 & 0.994 & 0.994 & 0.993 & 0.996 & 0.994 & 0.991 & 0.996 & 0.990 & 0.993 \\
        & T3 & 0.970 & 0.988 & 0.983 & 0.990 & 0.983 & 0.984 & 0.989 & 0.984 & 0.980 & 0.982 & 0.977 & 0.983 \\
        & L1 & 0.991 & 0.999 & 0.997 & 0.999 & 0.998 & 0.999 & 0.998 & 0.995 & 0.997 & 0.998 & 0.996 & 0.997 \\
        & L2 & 0.986 & 0.997 & 0.994 & 0.996 & 0.998 & 0.997 & 0.996 & 0.996 & 0.991 & 0.998 & 0.992 & 0.995 \\
        & L3 & 0.981 & 0.992 & 0.986 & 0.990 & 0.987 & 0.992 & 0.989 & 0.990 & 0.984 & 0.987 & 0.985 & 0.988 \\
        & TN1 & 0.990 & 0.997 & 0.997 & 0.998 & 0.996 & 0.998 & 0.998 & 0.996 & 0.997 & 0.999 & 0.995 & 0.996 \\
        & TN2 & 0.988 & 0.996 & 0.991 & 0.995 & 0.993 & 0.994 & 0.996 & 0.993 & 0.991 & 0.996 & 0.988 & 0.993 \\
        & TN3 & 0.972 & 0.989 & 0.984 & 0.991 & 0.983 & 0.986 & 0.991 & 0.984 & 0.980 & 0.984 & 0.976 & 0.984 \\
        & LN1 & 0.992 & 0.998 & 0.997 & 0.998 & 0.996 & 0.999 & 0.998 & 0.994 & 0.997 & 0.998 & 0.996 & 0.997 \\
        & LN2 & 0.989 & 0.997 & 0.994 & 0.995 & 0.996 & 0.996 & 0.997 & 0.995 & 0.991 & 0.997 & 0.991 & 0.994 \\
        & LN3 & 0.980 & 0.991 & 0.986 & 0.991 & 0.987 & 0.990 & 0.991 & 0.990 & 0.982 & 0.987 & 0.982 & 0.987 \\
        & TP1 & 0.977 & 0.991 & 0.984 & 0.986 & 0.981 & 0.985 & 0.988 & 0.982 & 0.980 & 0.988 & 0.978 & 0.984 \\
        & TP2 & 0.979 & 0.991 & 0.988 & 0.991 & 0.988 & 0.990 & 0.992 & 0.988 & 0.982 & 0.991 & 0.985 & 0.988 \\
        & TP3 & 0.973 & 0.988 & 0.986 & 0.984 & 0.980 & 0.983 & 0.984 & 0.982 & 0.978 & 0.985 & 0.975 & 0.982 \\
        & LP1 & 0.976 & 0.988 & 0.980 & 0.981 & 0.978 & 0.981 & 0.985 & 0.981 & 0.978 & 0.986 & 0.977 & 0.982 \\
        & LP2 & 0.977 & 0.989 & 0.984 & 0.988 & 0.986 & 0.988 & 0.990 & 0.986 & 0.980 & 0.990 & 0.982 & 0.986 \\
        & LP3 & 0.970 & 0.984 & 0.981 & 0.983 & 0.978 & 0.981 & 0.981 & 0.980 & 0.976 & 0.984 & 0.976 & 0.980 \\
      \hline
      \multirow{3}{*}{Morphological}
        & MS1 & 0.984 & 0.991 & 0.987 & 0.989 & 0.986 & 0.991 & 0.989 & 0.986 & 0.983 & 0.993 & 0.984 & 0.988 \\
        & MS2 & 0.979 & 0.990 & 0.990 & 0.989 & 0.989 & 0.991 & 0.991 & 0.987 & 0.984 & 0.990 & 0.985 & 0.988 \\
        & MS3 & 0.965 & 0.983 & 0.973 & 0.980 & 0.974 & 0.976 & 0.986 & 0.978 & 0.968 & 0.979 & 0.971 & 0.976 \\
      \hline
      \multirow{6}{*}{Syntactic}
        & P1 & 0.925 & 0.938 & 0.933 & 0.932 & 0.923 & 0.921 & 0.929 & 0.914 & 0.930 & 0.924 & 0.915 & 0.926 \\
        & P2 & 0.980 & 0.995 & 0.991 & 0.985 & 0.986 & 0.984 & 0.989 & 0.992 & 0.990 & 0.988 & 0.986 & 0.988 \\
        & P3 & 0.974 & 0.995 & 0.990 & 0.983 & 0.987 & 0.980 & 0.987 & 0.987 & 0.988 & 0.983 & 0.985 & 0.985 \\
        & D1 & 0.826 & 0.839 & 0.828 & 0.847 & 0.799 & 0.825 & 0.865 & 0.858 & 0.835 & 0.815 & 0.788 & 0.830 \\
        & D2 & 0.913 & 0.941 & 0.918 & 0.930 & 0.912 & 0.921 & 0.956 & 0.939 & 0.915 & 0.924 & 0.904 & 0.925 \\
        & D3 & 0.923 & 0.957 & 0.930 & 0.938 & 0.924 & 0.940 & 0.968 & 0.953 & 0.928 & 0.940 & 0.925 & 0.939 \\
      \hline
      \multirow{3}{*}{Statistical}
        & WL & 0.713 & 0.693 & 0.656 & 0.652 & 0.624 & 0.650 & 0.680 & 0.662 & 0.675 & 0.623 & 0.594 & 0.657 \\
        & SL & 0.554 & 0.715 & 0.722 & 0.765 & 0.733 & 0.631 & 0.720 & 0.723 & 0.680 & 0.704 & 0.713 & 0.696 \\
        & DD & 0.391 & 0.521 & 0.399 & 0.459 & 0.344 & 0.440 & 0.574 & 0.562 & 0.401 & 0.383 & 0.547 & 0.456 \\\hline
    \end{tabular}
    }
  \end{center}
\end{table}%

\begin{table}[htbp]
  \caption{As Table~\ref{tab:results_bin_toe}, but with \efcamdatbin\
  in place of \toeflbin.}
  \label{tab:results_bin_efc}
  \begin{center}
    \resizebox{\textwidth}{!} {
    \begin{tabular}{|c|c|c|c|c|c|c|c|c|c|c|c|c||c|}
      \cline{3-14}
      \multicolumn{2}{c|}{} & ARA & CHI & FRE & GER & HIN & ITA & JPN & KOR & RUS & SPA & TUR &Average \\\hline
      \multirow{18}{*}{Lexical}
        & T1 & 0.990 & 0.992 & 0.993 & 0.990 & 0.992 & 0.993 & 0.991 & 0.990 & 0.994 & 0.989 & 0.993 & 0.992 \\
        & T2 & 0.981 & 0.985 & 0.985 & 0.985 & 0.990 & 0.985 & 0.985 & 0.986 & 0.986 & 0.982 & 0.987 & 0.985 \\
        & T3 & 0.955 & 0.971 & 0.973 & 0.971 & 0.971 & 0.971 & 0.971 & 0.970 & 0.977 & 0.961 & 0.968 & 0.969 \\
        & L1 & 0.989 & 0.992 & 0.995 & 0.992 & 0.991 & 0.993 & 0.991 & 0.987 & 0.992 & 0.989 & 0.993 & 0.991 \\
        & L2 & 0.984 & 0.987 & 0.986 & 0.990 & 0.991 & 0.988 & 0.987 & 0.985 & 0.988 & 0.982 & 0.987 & 0.987 \\
        & L3 & 0.967 & 0.974 & 0.975 & 0.974 & 0.978 & 0.974 & 0.978 & 0.974 & 0.978 & 0.966 & 0.975 & 0.974 \\
        & TN1 & 0.987 & 0.990 & 0.990 & 0.990 & 0.992 & 0.990 & 0.988 & 0.988 & 0.992 & 0.988 & 0.991 & 0.990 \\
        & TN2 & 0.980 & 0.982 & 0.984 & 0.984 & 0.987 & 0.984 & 0.983 & 0.984 & 0.986 & 0.981 & 0.986 & 0.984 \\
        & TN3 & 0.958 & 0.974 & 0.974 & 0.972 & 0.972 & 0.970 & 0.969 & 0.971 & 0.977 & 0.960 & 0.970 & 0.970 \\
        & LN1 & 0.986 & 0.990 & 0.991 & 0.991 & 0.992 & 0.990 & 0.988 & 0.985 & 0.992 & 0.986 & 0.991 & 0.989 \\
        & LN2 & 0.983 & 0.986 & 0.985 & 0.989 & 0.988 & 0.986 & 0.985 & 0.983 & 0.987 & 0.981 & 0.986 & 0.985 \\
        & LN3 & 0.965 & 0.977 & 0.975 & 0.974 & 0.977 & 0.973 & 0.973 & 0.973 & 0.977 & 0.966 & 0.974 & 0.973 \\
        & TP1 & 0.973 & 0.979 & 0.979 & 0.971 & 0.980 & 0.980 & 0.975 & 0.975 & 0.981 & 0.966 & 0.981 & 0.976 \\
        & TP2 & 0.974 & 0.984 & 0.979 & 0.977 & 0.985 & 0.983 & 0.977 & 0.981 & 0.985 & 0.973 & 0.983 & 0.980 \\
        & TP3 & 0.961 & 0.978 & 0.972 & 0.968 & 0.977 & 0.973 & 0.972 & 0.974 & 0.977 & 0.960 & 0.977 & 0.972 \\
        & LP1 & 0.947 & 0.958 & 0.961 & 0.961 & 0.968 & 0.970 & 0.966 & 0.975 & 0.972 & 0.956 & 0.971 & 0.968 \\
        & LP2 & 0.948 & 0.954 & 0.962 & 0.967 & 0.965 & 0.963 & 0.966 & 0.971 & 0.972 & 0.963 & 0.972 & 0.972 \\
        & LP3 & 0.931 & 0.943 & 0.952 & 0.958 & 0.956 & 0.953 & 0.957 & 0.965 & 0.967 & 0.951 & 0.966 & 0.967 \\
      \hline
      \multirow{3}{*}{Morphological}
        & MS1 & 0.975 & 0.980 & 0.981 & 0.974 & 0.986 & 0.984 & 0.976 & 0.980 & 0.981 & 0.970 & 0.985 & 0.979 \\
        & MS2 & 0.974 & 0.982 & 0.980 & 0.982 & 0.987 & 0.980 & 0.979 & 0.982 & 0.983 & 0.973 & 0.985 & 0.981 \\
        & MS3 & 0.954 & 0.969 & 0.966 & 0.963 & 0.970 & 0.962 & 0.967 & 0.971 & 0.974 & 0.959 & 0.972 & 0.966 \\
      \hline
      \multirow{6}{*}{Syntactic}
        & P1 & 0.947 & 0.965 & 0.946 & 0.944 & 0.962 & 0.951 & 0.957 & 0.958 & 0.959 & 0.932 & 0.968 & 0.954 \\
        & P2 & 0.961 & 0.975 & 0.970 & 0.968 & 0.975 & 0.971 & 0.967 & 0.970 & 0.973 & 0.955 & 0.975 & 0.969 \\
        & P3 & 0.960 & 0.981 & 0.968 & 0.966 & 0.979 & 0.973 & 0.972 & 0.977 & 0.974 & 0.956 & 0.977 & 0.971 \\
        & D1 & 0.926 & 0.941 & 0.921 & 0.923 & 0.935 & 0.911 & 0.931 & 0.936 & 0.938 & 0.901 & 0.948 & 0.928 \\
        & D2 & 0.943 & 0.960 & 0.950 & 0.945 & 0.955 & 0.941 & 0.949 & 0.955 & 0.955 & 0.928 & 0.967 & 0.950 \\
        & D3 & 0.946 & 0.967 & 0.954 & 0.947 & 0.962 & 0.946 & 0.953 & 0.963 & 0.960 & 0.937 & 0.972 & 0.955 \\
      \hline
      \multirow{3}{*}{Statistical}
        & WL & 0.863 & 0.893 & 0.867 & 0.847 & 0.892 & 0.873 & 0.880 & 0.885 & 0.885 & 0.843 & 0.887 & 0.874 \\
        & SL & 0.861 & 0.888 & 0.876 & 0.878 & 0.888 & 0.854 & 0.900 & 0.910 & 0.905 & 0.839 & 0.922 & 0.884 \\
        & DD & 0.836 & 0.873 & 0.862 & 0.863 & 0.886 & 0.828 & 0.895 & 0.903 & 0.891 & 0.820 & 0.912 & 0.870 \\\hline
    \end{tabular}
    }
  \end{center}
\end{table}%

\begin{table}[htbp]
  \caption{As Table~\ref{tab:results_bin_toe}, but with \redditbin\ in
  place of \toeflbin.}
  \label{tab:results_bin_red}
  \begin{center}
    \resizebox{\textwidth}{!} {
    \begin{tabular}{|c|c|c|c|c|c|c|c|c|c|c|c|c||c|}
      \cline{3-14}
      \multicolumn{2}{c|}{} & FIN & FRE & GER & ITA & NED & NOR & POL & POR & ROM & SPA & SWE & Average \\\hline
      \multirow{18}{*}{Lexical}
                            & T1 & 0.764 & 0.784 & 0.786 & 0.758 & 0.739 & 0.749 & 0.774 & 0.774 & 0.771 & 0.757 & 0.758 & 0.765 \\
                            & T2 & 0.752 & 0.772 & 0.778 & 0.740 & 0.725 & 0.738 & 0.770 & 0.762 & 0.762 & 0.744 & 0.735 & 0.753 \\
                            & T3 & 0.701 & 0.710 & 0.725 & 0.688 & 0.671 & 0.685 & 0.716 & 0.701 & 0.704 & 0.681 & 0.674 & 0.696 \\
                            & L1 & 0.760 & 0.781 & 0.780 & 0.757 & 0.737 & 0.745 & 0.774 & 0.771 & 0.767 & 0.758 & 0.753 & 0.762 \\
                            & L2 & 0.758 & 0.774 & 0.780 & 0.739 & 0.726 & 0.741 & 0.768 & 0.760 & 0.765 & 0.744 & 0.739 & 0.754 \\
                            & L3 & 0.707 & 0.726 & 0.731 & 0.689 & 0.684 & 0.695 & 0.725 & 0.710 & 0.711 & 0.695 & 0.682 & 0.705 \\
                            & TN1 & 0.739 & 0.763 & 0.764 & 0.735 & 0.715 & 0.729 & 0.752 & 0.752 & 0.752 & 0.735 & 0.737 & 0.743 \\
                            & TN2 & 0.739 & 0.756 & 0.762 & 0.725 & 0.710 & 0.724 & 0.760 & 0.752 & 0.751 & 0.731 & 0.724 & 0.739 \\
                            & TN3 & 0.693 & 0.704 & 0.717 & 0.681 & 0.656 & 0.676 & 0.711 & 0.697 & 0.697 & 0.675 & 0.665 & 0.688 \\
                            & LN1 & 0.733 & 0.762 & 0.757 & 0.732 & 0.713 & 0.726 & 0.756 & 0.753 & 0.751 & 0.738 & 0.730 & 0.741 \\
                            & LN2 & 0.740 & 0.762 & 0.766 & 0.726 & 0.711 & 0.725 & 0.758 & 0.752 & 0.755 & 0.736 & 0.726 & 0.742 \\
                            & LN3 & 0.700 & 0.720 & 0.721 & 0.684 & 0.668 & 0.688 & 0.722 & 0.708 & 0.706 & 0.687 & 0.675 & 0.698 \\
                            & TP1 & 0.702 & 0.728 & 0.725 & 0.695 & 0.675 & 0.695 & 0.729 & 0.719 & 0.718 & 0.700 & 0.691 & 0.707 \\
                            & TP2 & 0.711 & 0.742 & 0.739 & 0.700 & 0.687 & 0.709 & 0.743 & 0.733 & 0.729 & 0.707 & 0.696 & 0.718 \\
                            & TP3 & 0.688 & 0.704 & 0.707 & 0.671 & 0.654 & 0.676 & 0.719 & 0.697 & 0.698 & 0.673 & 0.661 & 0.686 \\
                            & LP1 & 0.692 & 0.701 & 0.703 & 0.659 & 0.644 & 0.675 & 0.703 & 0.701 & 0.689 & 0.680 & 0.671 & 0.689 \\
                            & LP2 & 0.702 & 0.714 & 0.711 & 0.681 & 0.657 & 0.689 & 0.714 & 0.713 & 0.702 & 0.687 & 0.676 & 0.701 \\
                            & LP3 & 0.655 & 0.674 & 0.677 & 0.632 & 0.634 & 0.645 & 0.691 & 0.669 & 0.669 & 0.657 & 0.643 & 0.668 \\
      \hline
      \multirow{3}{*}{Morphological}
                            & MS1 & 0.706 & 0.733 & 0.727 & 0.701 & 0.681 & 0.695 & 0.724 & 0.724 & 0.720 & 0.699 & 0.692 & 0.709 \\
                            & MS2 & 0.705 & 0.732 & 0.739 & 0.699 & 0.681 & 0.704 & 0.735 & 0.726 & 0.723 & 0.698 & 0.691 & 0.712 \\
                            & MS3 & 0.674 & 0.691 & 0.699 & 0.663 & 0.644 & 0.666 & 0.706 & 0.683 & 0.685 & 0.661 & 0.651 & 0.675 \\
      \hline
      \multirow{6}{*}{Syntactic}
                            & P1 & 0.601 & 0.640 & 0.627 & 0.611 & 0.593 & 0.605 & 0.659 & 0.629 & 0.623 & 0.608 & 0.599 & 0.618 \\
                            & P2 & 0.638 & 0.677 & 0.663 & 0.642 & 0.622 & 0.643 & 0.688 & 0.678 & 0.665 & 0.645 & 0.627 & 0.653 \\
                            & P3 & 0.631 & 0.665 & 0.648 & 0.632 & 0.604 & 0.631 & 0.681 & 0.656 & 0.657 & 0.633 & 0.613 & 0.641 \\
                            & D1 & 0.608 & 0.617 & 0.609 & 0.597 & 0.586 & 0.598 & 0.657 & 0.607 & 0.617 & 0.607 & 0.588 & 0.608 \\
                            & D2 & 0.610 & 0.638 & 0.624 & 0.616 & 0.585 & 0.606 & 0.666 & 0.630 & 0.632 & 0.605 & 0.589 & 0.618 \\
                            & D3 & 0.603 & 0.629 & 0.618 & 0.602 & 0.578 & 0.608 & 0.646 & 0.616 & 0.621 & 0.605 & 0.585 & 0.610 \\
      \hline
      \multirow{3}{*}{Statistical}
                            & WL & 0.576 & 0.587 & 0.557 & 0.564 & 0.563 & 0.571 & 0.588 & 0.588 & 0.588 & 0.550 & 0.571 & 0.573 \\
                            & SL & 0.540 & 0.501 & 0.529 & 0.500 & 0.570 & 0.511 & 0.548 & 0.506 & 0.549 & 0.497 & 0.568 & 0.529 \\
                            & DD & 0.560 & 0.548 & 0.523 & 0.535 & 0.566 & 0.552 & 0.580 & 0.550 & 0.554 & 0.543 & 0.566 & 0.552 \\\hline
    \end{tabular}
    }
  \end{center}
\end{table}%



\subsection{Results of the binary experiments: Predicting if the
speaker is native or non-native}
\label{sec:binaryresults}

Tables~\ref{tab:results_bin_toe}, \ref{tab:results_bin_efc},
\ref{tab:results_bin_red} show how each feature set performs in the
L1-vs-EN binary tasks on the different datasets used. Compared to the
multiclass NLI task, accuracy is much higher for every feature set on
every dataset. This time, most feature sets behave
well, especially \toeflbin\ and \efcamdatbin, except for statistical
features, which perform worse than the others on
all the datasets.

Accuracy on the \toeflbin\ and \efcamdatbin\ corpora is very high in
many cases, especially for lexical features, which often give rise to
accuracy values in the 0.96--0.99 range. In \efcamdatbin\ even the
statistical features (WL, SL, DD) give rise to high accuracy values.
We conjecture this to be caused by the differences in context and
motivations that underlie the \locness\ documents with respect to the
\toefl\ and \efcamdat\ documents, which makes the task of separating
\locness\ from the other two easier.

Conversely, \redditbin\ uses \strikeout{instead} the same source for
non-native and native documents, thus factoring out the aspects that
make L1-vs-EN classification easier for the other two
datasets. Accuracy values are thus lower than in \toeflbin\ and
\efcamdatbin, while still very good. Here, the statistical features
have accuracy scores that are close to those of the random classifier
(whose expected accuracy is 0.5), indicating that there are hardly any
significant differences in the phenomena they represent (i.e., word
length, sentence length, and dependency depth) between native
production and non-native production, and that any clue used by the
learning algorithm to make a correct prediction is based on lexical /
morphological / syntactic features.

It is relevant to note that, for all corpora, no L1 emerges as
significantly harder or easier to \replace{recognize}{recognise} with
respect to the others, and that no L1 shows a different trend in the
relative accuracy scored by the different types of features.

\subsection{Feature analysis: Lexical, morphological, and syntactic
features}
\label{sec:lexmorsyn}

\noindent In order to show the power of SVM-based explainable machine
learning for characterising language transfer, we analyse the results
of multiclass classification for NLI, and we draw our examples from
two sample L1s, Spanish and Italian, as emerging from two sample
datasets, \efcamdat\ and \toefl. (We concentrate on \efcamdat\ and
\toefl\ because they are the two datasets on which the best accuracy
is reached, so the intuitions about feature importance that we can
draw from them are more reliable.) Similar analyses can be carried out
on other L1s, on other datasets, and on the L1-vs-EN task.

In order to \replace{gain insights}{gain insight} into L1-specific
patterns, we look at the features that the machine deemed most
discriminant for each language group. 
In order to do this, we inspect the parameter values (hereafter:
``coefficients'') that the SVM assigned to each feature in the ``All''
experiment reported in Table~\ref{tab:results_L1_sets}; as explained
in Section~\ref{sec:NLIandSLA}, a coefficient determines how the value
of the corresponding feature's relative frequency in a document
contributes to the classification decision for that document, with
coefficients of high absolute magnitude indicating a large impact on
the decision, and with the sign of the coefficient indicating whether
this value weighs towards assigning ($+$) or not assigning ($-$) the
corresponding L1.  In other words, positive coefficients identify
overuse patterns, whilst negative coefficients identify underuse
patterns.

It must be pointed out that overuse and underuse patterns emerge from
a \emph{contrastive} L2-based perspective, i.e., by comparing the
written output of one linguistic group to that of all the other
linguistic groups. As a consequence, the linguistic behaviour of an L1
group must be viewed as a relative rather than an absolute
phenomenon. What can be observed are indeed discriminant linguistic
deviations that characterise specific L1 groups only with reference to
the other L1 groups.
Such deviations can occasionally coincide, but not necessarily, with
errors (e.g., spelling mistakes typical of one particular L1 group).


\subsubsection{Case Study 1: Spanish as L1 in \efcamdat.}
\label{sec:spanishinefcamdat}

\noindent In \efcamdat, the coefficients with the highest absolute
magnitude turn out to correspond to named entities. In view of what we
said in Section~\ref{sec:lexicalfeatures}, this is unsurprising, since
many essays of which \efcamdat\ consists of deal, as discussed in
Section~\ref{sec:multiclass}, with everyday experiences of the
speakers; as such, they are likely to contain many named entities that
refer to the local culture / environment of the speaker, and that thus
``give away'' the nationality of the speaker. However, named entities
are uninteresting to our goals, and we thus do not discuss them; we
thus discuss features other than named entities, starting with lexical
features that play an important role for specific L1s.

In the case of Spanish learners, the machine identified two
particularly discriminant features in \efcamdat, i.e., the use of the
words *\emph{de} (coefficient:\ $+$2.74) and *\emph{diferent}
(coefficient:\ $+$2.49).\footnote{As standard in the linguistic
literature, we prefix with a star (*) all incorrect uses of English.}
By examining some usage examples (e.g., \emph{de train} or \emph{de
evening}), we noticed that the former is a misspelling of the
determiner \emph{the}, likely influenced by the phonology of
Spanish. Indeed, since Spanish does not exhibit the phoneme
/\dh/,\footnote{The sound /\dh/ exists in certain areas of the
Spanish-speaking countries but only as an allophone.} Spanish learners
of English approximate the voiced interdental fricative to the dental
sound /d/, which, conversely, belongs in the Spanish phonological
system. Interestingly, voice prevails over manner and place of
articulation. In fact, Spanish does make use of the phoneme
/\texttheta/, which is the unvoiced counterpart of /\dh/. Yet,
learners instinctively approximate the latter to /d/.

As to the misspelling of the adjective \emph{different}, it appears to
follow the spelling of the Spanish equivalent \emph{diferente}, in
which the double consonant \emph{ff} is reduced to a single consonant,
following its pronunciation. It must be noted that, amongst the
observed examples involving the term \emph{different}, instances of
pluralisation of the adjective can be detected (e.g., *\emph{diferents
areas} or *\emph{diferents types}), brought about by the agreement
rules of Spanish.

Although Spanish learners exhibit a distinct tendency to start a new
sentence with the adverb \emph{never} (coefficient:\ $+$2.42) more
often than learners from other L1 backgrounds, e.g., \emph{never
forget your family} (...) or \emph{never the ball must touch} (...),
this fact cannot be directly linked to language transfer. On the
contrary, overuse of the bigram \emph{because is} (coefficient:\
$+$1.97) as in the examples \emph{because is ugly} or \emph{because is
very strange} point to the omission of the dummy pronoun, typical of
the Spanish language.

In terms of underuse patterns, the most relevant habits captured by
the machine concern the beginning of new sentences. The machine
assigned the highest coefficients to the bigrams ``\emph{. so}''
(coefficient:\ $-$2.34), ``. \emph{but}'' (coefficient:\ $-$1.74), and
``. \emph{and}'' (coefficient:\ $-$1.52), a sign that Spanish learners
tend to avoid starting a new sentence with \emph{so}, \emph{but} and
\emph{and}, just as they would do in their mother tongue.


\subsubsection{Case Study 2: Italian as L1 in \toefl.}
\label{sec:italianintoefl}

\noindent Differently from the case study discussed in
Section~\ref{sec:spanishinefcamdat}, here the coefficients with the
highest absolute magnitude do not belong to named entities, for the
simple reason that, as mentioned in Section~\ref{sec:NLIresults},
named entities were masked off from \toefl\ directly by its creators.

The most relevant feature marking the production of Italian learners
in the \toefl\ dataset is the trigram \emph{I think that} preceded by
punctuation or by another word\emph{.} The pattern appears with a
coefficient of $+$3.84. Its presence can be accounted for by the
nature of the TOEFL exam, in which students are often prompted to give
their opinion on a variety of topics. However, if its appearance were
the sole result of following the writing instructions, it should be
evenly distributed across the different L1s, and therefore lose its
importance as a feature. Rather, its high incidence in the writings of
Italian learners vouches for its significance. Consistent resort to
the above-mentioned expression could indeed
stem from the convenient and safe equivalence between the Italian
\emph{penso che} and the English \emph{I think that}. A further
hypothesis could be that Italian speakers simply employ (in their
mother tongue) the phrase \emph{penso che} with greater frequency
compared to other L1 speakers. Yet, evidence from comparable corpora
should be provided to support this thesis.

The second most discriminant feature in the \toefl\ dataset is the
bigram ``token + :'' to which the machine assigned a coefficient of
+3.71. The use of the colon in English is indeed problematic for
Italian learners. In fact, in Italian, the colon can be either used to
introduce a list of items or to illustrate a concept, whilst in
English only the former is allowed. By looking into the occurrences of
``token + :'', we were able to identify many examples of misuse (e.g.,
\emph{one thing seems clear: long term scars} or \emph{you take a
risk: you change your type of}), which can account for the high number
of occurrences of the pattern. 

Typical of Italian learners is the trigram ``. \emph{in fact}''
(coefficient:\ $+$2.98) together with its misspelt alternative form
*\emph{infact} (coefficient:\ $+$2.94). The latter can also be found
in the \efcamdat\ dataset with a coefficient of $+$2.87. In Italian,
\emph{infatti} is a highly occurring word. Nevertheless, despite a
superficial resemblance, \emph{in fact} and \emph{infatti} have
different meanings and functions. The former provides more detailed
information about a topic previously introduced. The latter confirms
something that was previously stated by means of a causal
relationship. Examples of misuse in \toefl\ are \emph{he or} \emph{she
takes risks. In fact, in the business world if} (...) or \emph{I think
in fact that a specialised worker has} (...).

Particularly distinctive in \toefl\ (coefficient:\ $+$2.69) are
reporting verbs (\emph{believe, decide,} and \emph{think}) followed by
a \emph{that}-clause and governing a subject.

From a strictly lexical point of view, two lemmas are very relevant in
\toefl\ as they mark the production of Italian learners: the adverb
\emph{probably} (coefficient:\ $+$2.59) and the noun
\emph{possibility} (coefficient:\ $+$2.45). The former is occasionally
misplaced in a way that can be traced back to Italian syntax, e.g.,
\emph{a question that probably I will present to myself}.

As regards patterns of underuse, the most discriminating feature is
the verb \emph{get} which exhibits a coefficient of $-$3.07. Indeed,
compared to other L1 speakers of English, Italians are reluctant to
use such a verb. Moreover, they seem to rarely start a new sentence
with a conjunction, a fact that can be accounted for by the stylistic
rules of written Italian, which do not allow for such a pattern.


\subsection{Feature analysis: Statistical features}
\label{sec:Statistical}

\noindent
Although statistical features (WL, SL, DD) do not seem to be very
helpful in discriminating the different L1s, they can nonetheless be
investigated in order to gather information concerning the linguistic
habits of speakers of English from different linguistic
backgrounds. In this section we compare the values taken by the
statistical features with respect to two orthogonal dimensions, i.e.,
the L1 and the proficiency level.

\newcolumntype{C}{>{\centering\arraybackslash}p{0.12\textwidth}|}

\begin{table}[htbp]
  \caption{Average values of the statistical features for each
  language group on the \efcamdat\ dataset.}
  \label{tab:statistical}
  \begin{center}
    \begin{tabular}{|c||CCC}
      \hline
      \multirow{2}{*}{} & avg WL & avg SL & avg DD \\
      \hline\hline
      ARA & 3.76 & 11.64 & 2.16 \\
      CHI & 3.76 & 11.86 & 2.10 \\
      FRE & 3.82 & 12.59 & 2.21 \\
      GER & 3.94 & 13.18 & 2.29 \\
      HIN & 3.77 & 11.68 & 2.09 \\
      ITA & 3.89 & 14.11 & 2.41 \\
      JPN & 3.80 & 11.18 & 2.03 \\
      KOR & 3.70 & 10.61 & 1.95 \\
      SPA & 3.84 & 13.12 & 2.26 \\ 
      RUS & 3.87 & 11.91 & 2.19 \\
      TUR & 3.75 & 10.37 & 1.94 \\
      \hline
      (All) & 3.83 & 12.13 & 2.17 \\
      \hline
    \end{tabular}
  \end{center}
\end{table}%


Table~\ref{tab:statistical} shows information on the average values of
the statistical features for each language group on the \efcamdat\
dataset.\footnote{Similar trends are observed on the other datasets.}
German learners have a slightly higher word length average than other
L1s, with Italians in second place. Italian, German, Spanish, and
French learners produce longer sentences if compared to other L1s.
Conversely, Turkish and Korean learners tend to be more succinct, with
the shorter sentence length (almost four words of difference between
Italian and Korean) also reflected by a smaller depth in the parse
tree.

\efcamdat\ essays are rated by 16 proficiency levels, 1 being the
lowest and 16 being the highest. We grouped essays by three ranges of
proficiency levels, thus deriving three subsets of the original
\efcamdat: low (\efcamdat-G1), containing essay in levels 1 to 5
included; intermediate (\efcamdat-G2), for levels 6 to 12; and
advanced (\efcamdat-G3), for levels 13 and higher. We compare the
frequency distribution of the statistical features across these three
proficiency groups.

\begin{figure}[t]
  \begin{center}
    \includegraphics[width=.80\textwidth]{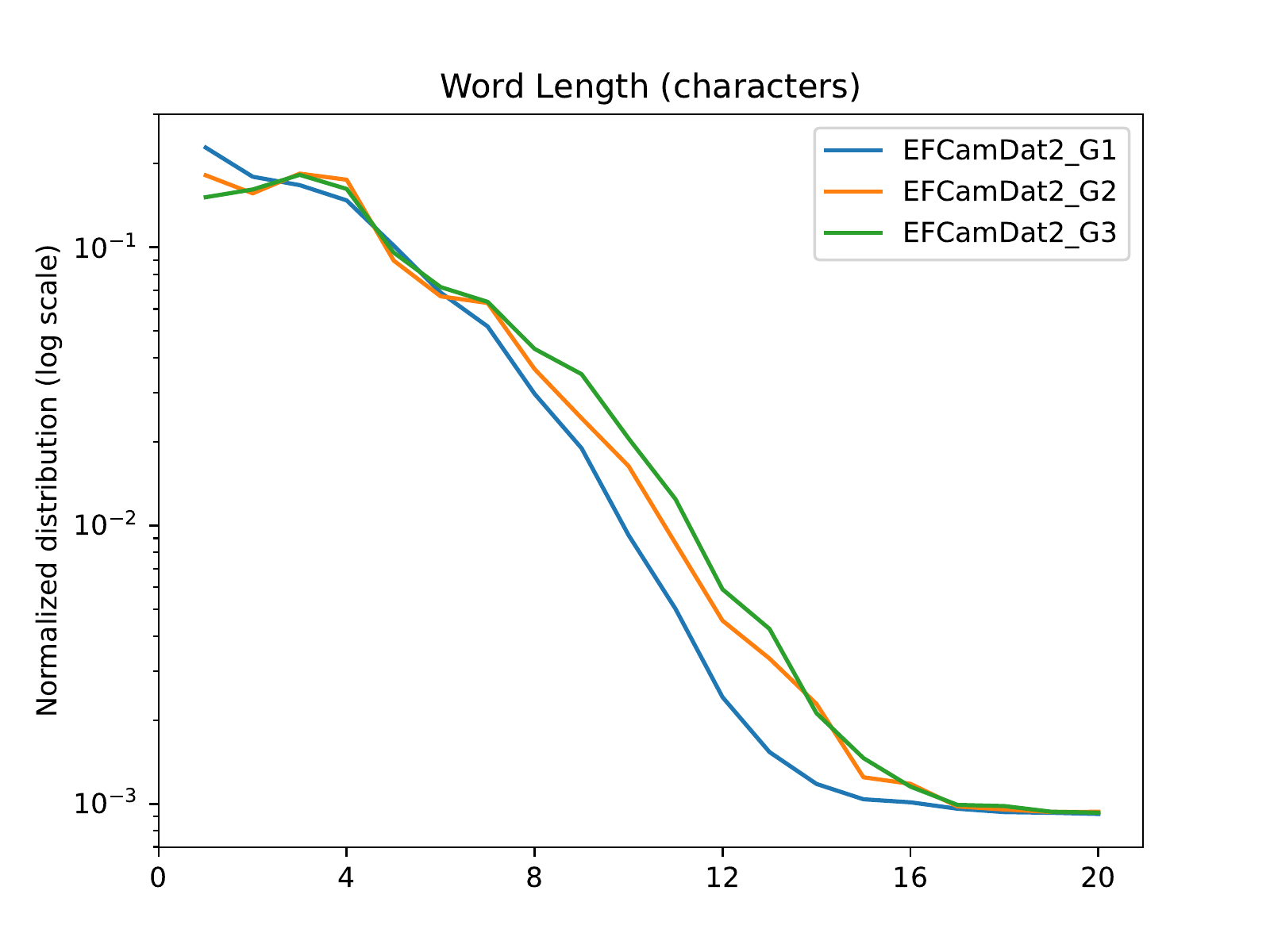}
    \caption{Comparison of normalized distribution of word length
    across proficiency groups in \efcamdat.}
    \label{fig:wordlength}
  \end{center}
\end{figure}

If we look at word length (Figure~\ref{fig:wordlength}) we can observe
that it increases at the increase of the proficiency level (a shift
towards right, i.e., towards longer words, of the curve of the
distribution).  (Average word length for documents in \efcamdat-G1 is
3.49, for \efcamdat-G2 is 3.85, and for \efcamdat-G3 is 4.12.) This is
unsurprising because longer words are, on average, less common in
language use, and the use of longer words denotes higher
sophistication in a learner's active vocabulary.

\begin{figure}[t]
  \begin{center}
    \includegraphics[width=.80\textwidth]{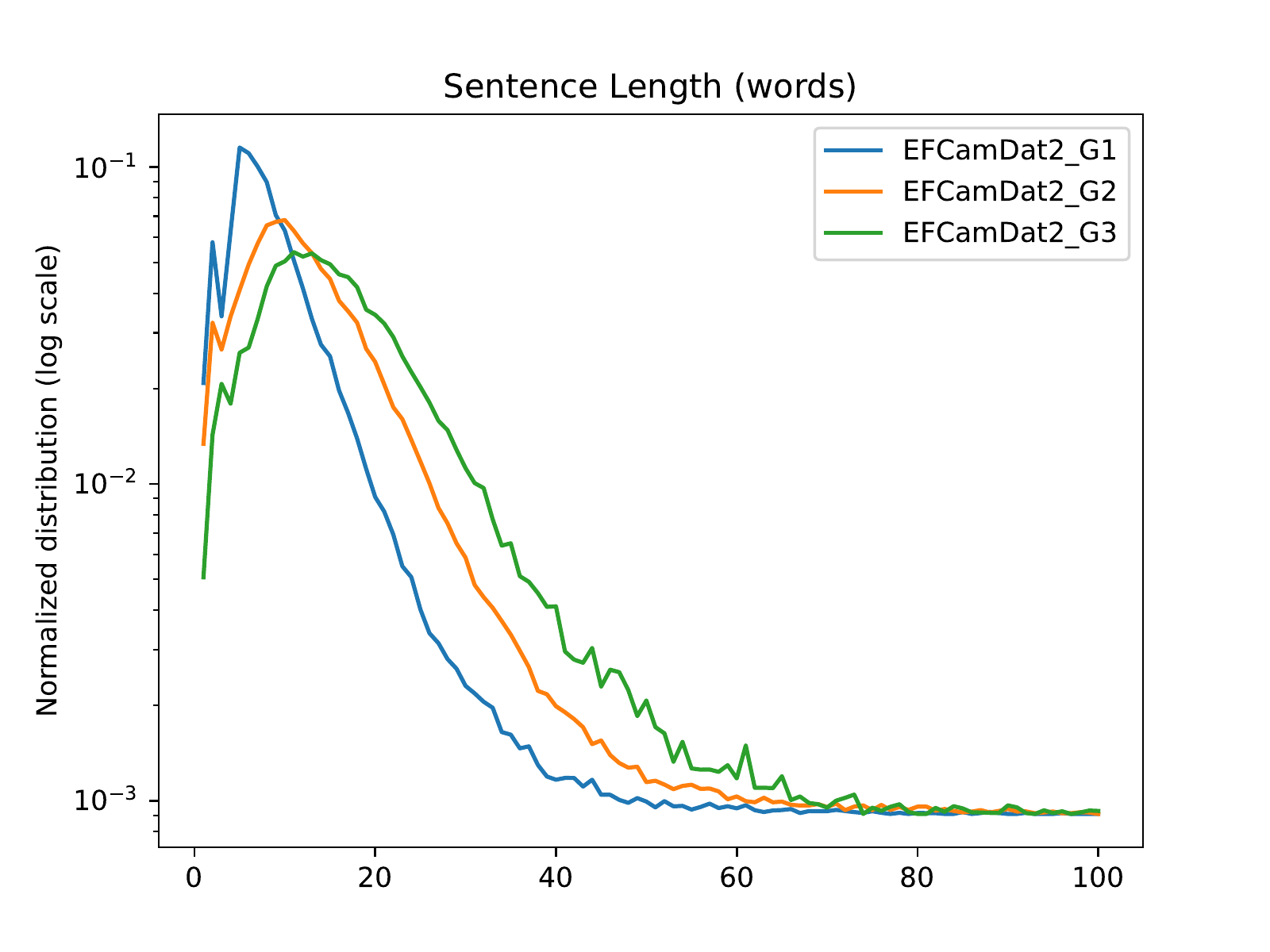}
    \caption{Comparison of normalized distribution of sentence length
    across proficiency groups in \efcamdat.}
    \label{fig:sent_length}
  \end{center}
\end{figure}

Sentence length (Figure~\ref{fig:sent_length}) follows a similar
pattern, with longer sentences being produced, on average, by more
proficient learners. Learners with a smaller command of their L2 tend
to produce shorter sentences, composed on average of 8.9 words; this
number doubles for the top proficiency group \efcamdat-G3, with an
average length of 16.7 words, while group \efcamdat-G2 places almost
exactly in the middle, with an average sentence length of 12.8 words.

\begin{figure}[t]
  \begin{center}
    \includegraphics[width=.80\textwidth]{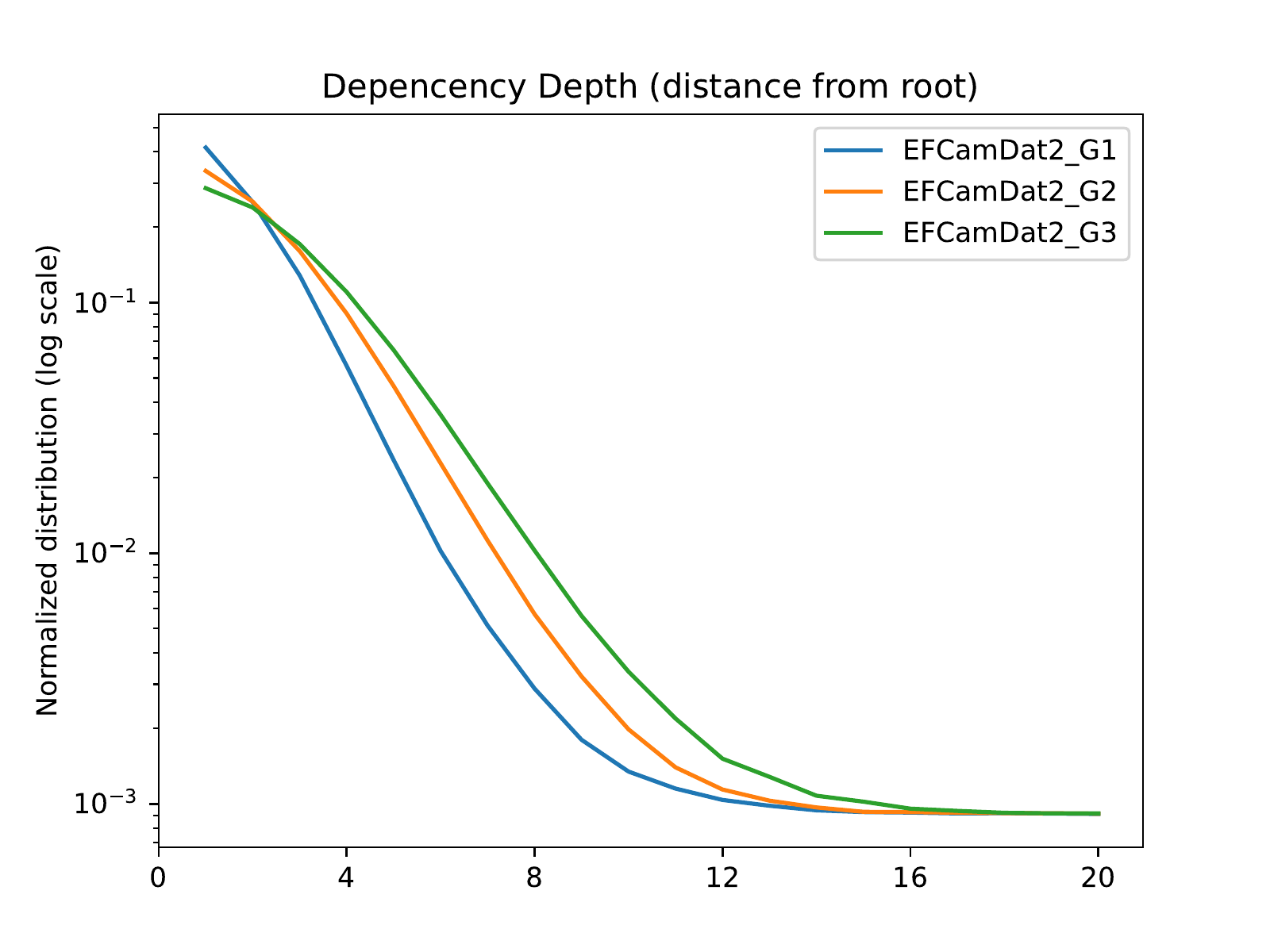}
    \caption{\label{fig:dependency}Comparison of normalized
    distribution of dependency depth (i.e., distance from root of the
    dependency tree) across proficiency groups in \efcamdat.}
  \end{center}
\end{figure}

The length of the sentence is obviously correlated with its
complexity, and thus with the depth of the dependency tree. This is
confirmed by the comparison of the frequency distributions of feature
DD (Figure~\ref{fig:dependency}): the lower proficiency group contains
sentences with a more shallow structure (average depth:\ 1.75), and
depth increases as proficiency improves (average depth:\ 2.18 for
\efcamdat-G2 and 2.57 for \efcamdat-G3). All in all, this is also
unsurprising, since more proficient learners have a higher command of
the syntax of the L2, and thus venture more often into more complex
syntactic structures and, as a consequence, into longer sentences.

\section{Conclusion}
\label{sec:Conclusion}

\noindent Explainable machine learning (EML) can be a very powerful
tool to investigate second language acquisition, and language transfer
in particular, especially when sizeable amounts of learner data for a
variety of different languages are available. We have shown how
interesting facts about language transfer emerge from the analysis of
the parameters of classifiers trained to perform native language
identification or native vs.\ non-native classification. The
classifiers we have discussed in this paper were trained via support
vector machines, but also other classifier-learning methods, such as
logistic regression, produce similarly interpretable classifiers. Each
parameter of the SVM classifier is associated to a feature, i.e., a
linguistic trait whose frequency of occurrence in the different
classes of interest (i.e., native speakers, non-native speakers,
non-native speakers of a specific L1) we want to exploit in order to
perform classification. Features to which the learning algorithm has
associated a value of high absolute magnitude represent linguistic
traits whose usage patterns significantly differ across the classes of
interest, with a positive value weighing towards assigning the class
and a negative value weighing against assigning it. We have shown, by
drawing examples from two among the classes we have investigated
(Spanish learners of English and Italian learners of English) how the
parameters that are assigned the values with the highest magnitude are
indeed associated with linguistic traits that are well-known to
characterize the linguistic production of those speakers. This shows
that performing native language identification, or native vs.\
non-native classification, via an EML algorithm, can be a valuable
tool for the scholar who investigates second language acquisition and
language transfer.

Where could improvements to these results come from? One promising
line of research could involve new EML methods. While until a few
years ago it was generally accepted that ML algorithms could generate
``black box'' (i.e., hardly inspectable) classifiers, the push towards
EML has increased in the last ten years, due to the fact that ML
algorithms are more and more frequently applied to high-stakes domains
(e.g., algorithms that decide if a convict should be granted parole,
algorithms that decide if a loan application should be considered
favourably or not, etc.), and that their decisions cannot be accepted
without an accompanying justification. Unfortunately, most research on
EML so far has targeted structured (i.e., tabular) data, and many
proposed solutions are hardly applicable to textual data because of
the high dimensionality of the latter. More research on EML for text
is needed, and this would hold promise for our application context.

Another factor of key importance for research on language transfer is
the quality of the available datasets. First, datasets of higher
quality than the ones we have used here could deliver more accurate
classifiers, which would allow drawing more reliable intuitions about
language transfer. An important step towards higher-quality datasets
would derive from having data in which the mother tongue of the
speaker is explicit; in the datasets we have used we can only
\emph{estimate} the speaker's L1 from its nationality / country of
residence, but these latter attributes are not always in a 1-to-1
correspondence with mother tongue, so it is not clear how many of the
inaccurate decisions that today's classifier return are due to
mislabelled training documents or mislabelled test documents. Another
important improvement might come from having better quality native
vs.\ non-native datasets than \toeflbin\ and \efcamdatbin. These
latter derive from the union of two datasets (a native dataset and a
non-native dataset) consisting of two different types of text, which
makes the binary classification task easier than it should be; the
availability of more homogeneous native vs.\ non-native datasets (such
as \redditbin) would thus be an important step towards addressing the
(still under-researched problem) of native vs.\ non-native
classification.

\section*{Acknowledgments}
\label{sec:acks}

\noindent The work by the 2nd author and 3rd author has been supported
by the \textsc{SoBigData++} project, funded by the European Commission
(Grant 871042) under the H2020 Programme INFRAIA-2019-1, and by the
\textsc{AI4Media} project, funded by the European Commission (Grant
951911) under the H2020 Programme ICT-48-2020. The authors' opinions
do not necessarily reflect those of the European Commission.


\theendnotes




\end{document}